\documentclass[lettersize,journal]{IEEEtran}
\usepackage{amsmath,amsfonts}
\usepackage{algorithmic}
\usepackage{algorithm}
\usepackage{array}
\usepackage[caption=false,font=normalsize,labelfont=sf,textfont=sf]{subfig}
\usepackage{textcomp}
\usepackage{stfloats}
\usepackage{url}
\usepackage{verbatim}
\usepackage{graphicx}
\usepackage{cite}
\usepackage{multirow}
\usepackage{color}
\usepackage[switch]{lineno}
\usepackage{subfig}
\usepackage{diagbox}

\begin{document}

\title{Dual-Perspective Semantic-Aware Representation Blending for Multi-Label Image Recognition with Partial Labels}

\author{Tao Pu, Tianshui Chen, Hefeng Wu, Yukai Shi, Zhijing Yang, Liang Lin
\IEEEcompsocitemizethanks{ \IEEEcompsocthanksitem Tao Pu, Hefeng Wu, and Liang Lin are with Sun Yat-Sen University, Guangzhou, China. Tianshui Chen, Yukai Shi and Zhijing Yang are with The Guangdong University of Technology, Guangzhou, China.\protect\\}
}

\markboth{Journal of \LaTeX\ Class Files,~Vol.~14, No.~8, August~2021}%
{Shell \MakeLowercase{\tiextit{et al.}}: A Sample Article Using IEEEtran.cls for IEEE Journals}


\maketitle


\begin{abstract}
Despite achieving impressive progress, current multi-label image recognition (MLR) algorithms heavily depend on large-scale datasets with complete labels, making collecting large-scale datasets extremely time-consuming and labor-intensive. Training the multi-label image recognition models with partial labels (MLR-PL) is an alternative way, in which merely some labels are known while others are unknown for each image. However, current MLP-PL algorithms rely on pre-trained image similarity models or iteratively updating the image classification models to generate pseudo labels for the unknown labels. Thus, they depend on a certain amount of annotations and inevitably suffer from obvious performance drops, especially when the known label proportion is low. To address this dilemma, we propose a dual-perspective semantic-aware representation blending (DSRB) that blends multi-granularity category-specific semantic representation across different images, from instance and prototype perspective respectively, to transfer information of known labels to complement unknown labels. Specifically, an instance-perspective representation blending (IPRB) module is designed to blend the representations of the known labels in an image with the representations of the corresponding unknown labels in another image to complement these unknown labels. Meanwhile, a prototype-perspective representation blending (PPRB) module is introduced to learn more stable representation prototypes for each category and blends the representation of unknown labels with the prototypes of corresponding labels, in a location-sensitive manner, to complement these unknown labels. Extensive experiments on the MS-COCO, Visual Genome, and Pascal VOC 2007 datasets show that the proposed DSRB consistently outperforms current state-of-the-art algorithms on all known label proportion settings. 
\end{abstract}


\section{Introduction} \label{sec:intro}
\begin{figure}[!t] 
  \centering
  \includegraphics[width=0.95\linewidth]{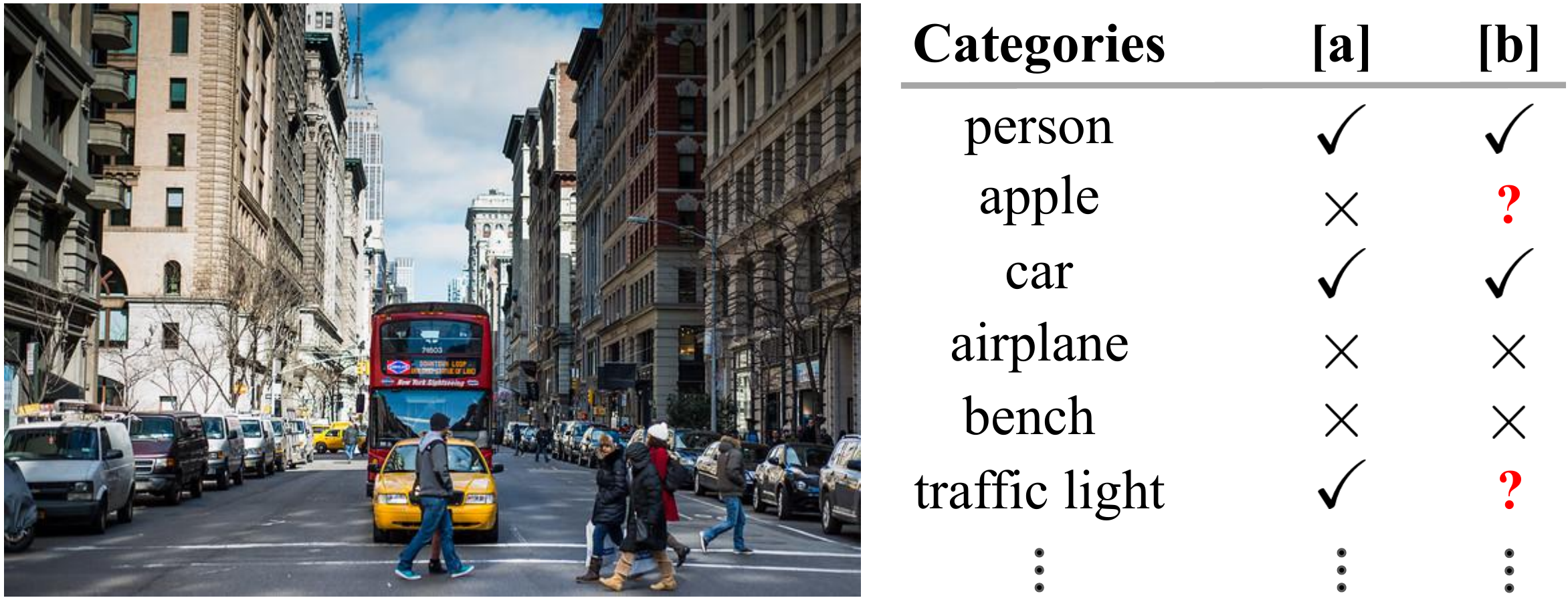}
  \caption{An example of MLR image with  [a] complete labels and [b] partial labels. $\checkmark$ denotes the corresponding category exists meanwhile $\times$ denotes it does not exist. In partial labels, {\color{red} $\textbf{?}$} denotes the original label is missing or the annotator does not know whether the corresponding category exists.}     
  \label{fig:task}     
\end{figure}

Compared with the single-label counterpart, multi-label image recognition (MLR) is a more practical and challenging task because the daily images inherently contain multiple semantic objects of diverse categories, which aims to find out all semantic labels from the input image. Recently, lots of efforts \cite{sun2014multi, gao2021learning, Li2021T-CYB, Zhang2020T-CYB, Liu2017T-CYB} are dedicated to addressing this task as it supports plenty of downstream applications of image content understanding ranging from recommendation systems \cite{Carrillo2013Multi, Zheng2014Context, Fu2022T-CYB} to content-based image retrieval \cite{Li2010Technique, Zhang2021Instance, lai2016instance, shen2021deep_tmm}. However, current leading algorithms \cite{Wu2020AdaHGNN, Chen2022KGGR, Chen2021P-GCN} heavily depend on large-scale datasets that have clean and complete labels, leading to the heavy dependencies on the time-consuming and labor-intensive annotation progress. To alleviate this dilemma, more recent works \cite{Durand2019CVPR, Chen2022SST, Pu2022SARB, Pu2022MLR-PPL} are dedicated to the more practical multi-label image recognition with partial labels (MLR-PL) task, in which merely a few positive and negative labels are provided whereas other labels are unknown, as shown in Figure \ref{fig:task}. In this way, it can dramatically reduce the annotation cost because annotating partial labels for each image is easier and more scalable than annotating complete labels.

Previous works \cite{Sun2017ICCV, Joulin2016ECCV} simply ignore the unknown labels or treat them as negative and adapt traditional MLR algorithms to address the MLR-PL task. Some other works \cite{Kim2022LargeLoss, Pu2022MLR-PPL} further propose to re-weight the loss of each sample and thus alleviate the effect of label noise. However, such adaptation may hurt the training process because it either loses some annotations or even incurs some incorrect labels, inevitably leading to obvious performance degradation. More recent works \cite{Durand2019CVPR, Huynh2020CVPR} introduce the pre-trained image similarity models or directly utilize the poor-performance models trained on partial labels to generate pseudo labels for the unknown labels, which can retrieve some unknown labels to alleviate this dilemma. Despite acknowledged progress, these algorithms rely on sufficient annotations and suffer from obvious performance drops, especially when the known label proportion is low, e.g., with merely 10\% known labels.

\begin{figure}[!t] 
  \centering
  \includegraphics[width=0.95\linewidth]{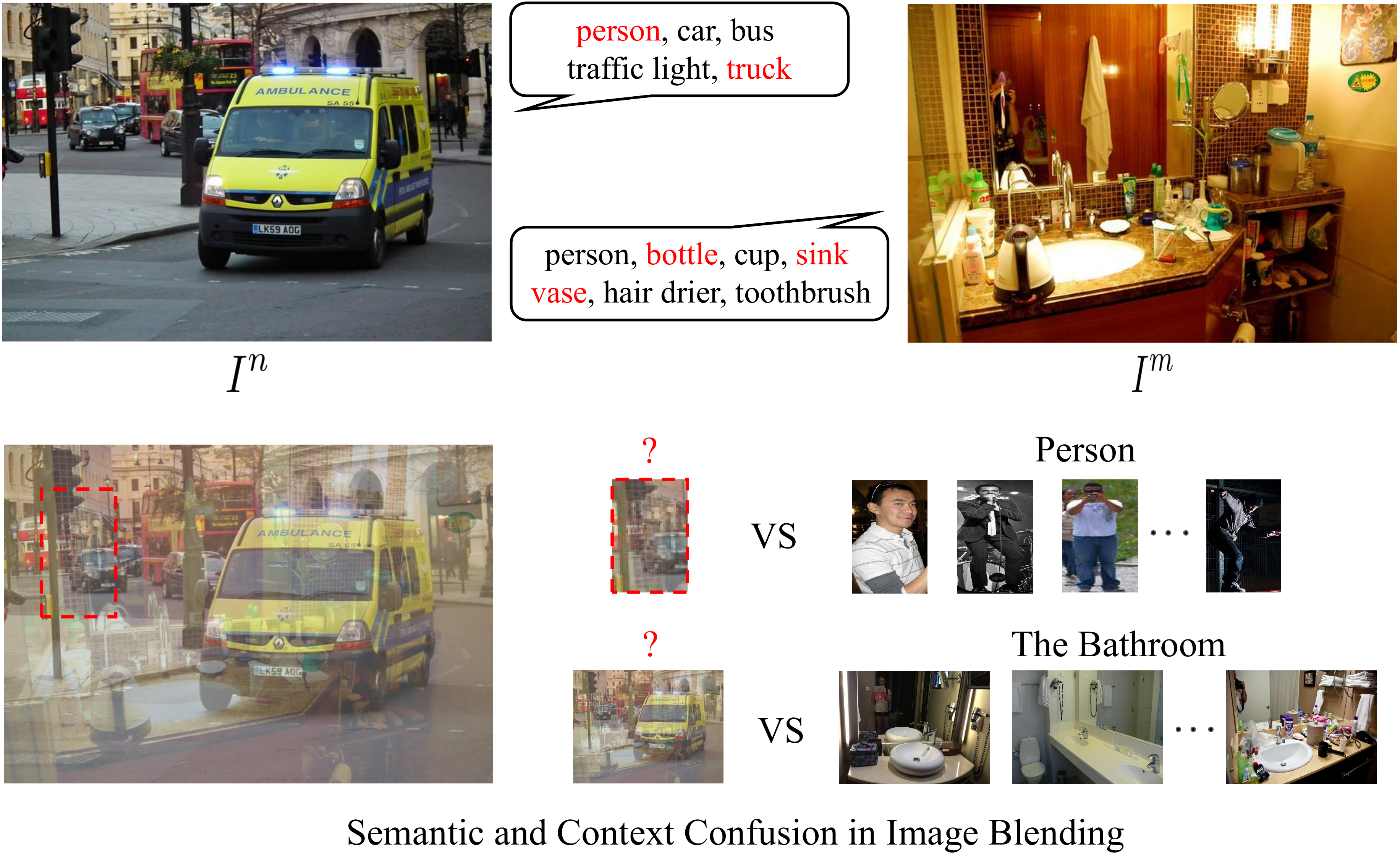}
  \caption{Two examples of images with partial labels (unknown labels are highlighted in red) and examples of semantic and context confusion in naive image blending.} 
  \label{fig:motivation}     
\end{figure}

Fortunately, a specific label $c$ that is unknown in one image $I^n$ may be known in another image $I^m$. Therefore, blending the information of known label $c$ from image $I^m$ to image $I^n$ may help to complement the unknown label $c$ for image $I^n$. However, blending two images via naive mixup operation \cite{Zhang2017Mixup} can hardly help facilitate the MLR-PL task since such operation may lead to semantic and context confusion, as discussed in the following. First, multi-label images inherently contain multiple objects of diverse semantic categories, which have variant sizes and scatters over the whole image. Simply blending two images may mix two objects with totally different semantics together, which semantically confuses the model during training. As shown in Figure \ref{fig:motivation}, blending image $I^m$ to image $I^n$ mixes the ``person" in $I^m$ and the ``traffic light" in $I^n$ together, which generate jumbly regions and may hurt the training process. Second, multi-label images contain multiple objects with heavy context dependencies, which can provide extra guidance to facilitate multi-label recognition \cite{Chen2019SSGRL,Chen2022KGGR}. Blending two images, especially if they belong to different scenes, may destroy these dependencies. Given two images captured in ``the street" and ``the bathroom" in Figure \ref{fig:motivation}, the scene context of ``the street" may provide confusing context information for recognizing the ``hair drier", ``toothbrush". In this work, we explore category-specific representation learning and blending to avoid the above-mentioned semantic and context confusion dilemma, which can better complement the unknown labels to facilitate the MLR-PL task. It does not rely on sufficient annotations, and thus it obtains consistently better performance on different know label proportion settings.

Specifically, we propose a semantic-aware representation blending (DSRB) framework, which learns category-specific representation for each image and then performs category-specific representation blending to complement the unknown labels. It consists of two crucial modules that blend category-specific representation from instance and prototype perspectives, respectively. The DSRB framework builds on a category-specific representation learning (CSRL) module that incorporates category semantics to guide learning category-specific semantic representation. Then, an instance-perspective representation blending (IPRB) module is designed to blend the representation of the known label $c$ in one image $I^m$ to the representation of the corresponding unknown label $c$ in another image $I^n$, and thus it can complement the unknows label $c$ for image $I^n$. Meanwhile, a prototype-perspective representation blending (PPRB) module is proposed to learn more robust representation prototypes for each category and blends the representation of unknown labels with the prototypes of corresponding labels, in a location-sensitive manner, to complement these unknown labels. In this way, we can simultaneously generate diverse and stable blended representations to complement the unknown labels and thus facilitate the MLR-PL task.

A preliminary version of this work was presented as a conference paper \cite{Pu2022SARB}. In this version, we inherit the idea of semantic-aware representation blending and strengthen this work from three aspects in the following. First, the conference version directly blends the representation vector, which does not consider the object location information and may lead to semantic confusion. In this work, we introduce location-sensitive representation blending on the feature maps and thus avoid generating confusing training samples. Second, we present more in-depth analyses on image blending for multi-label images, which can better explain the motivation of the DSRB framework. Finally, more in-depth discussions and experimental analyses have been added to demonstrate our motivations and evaluate the actual contribution of each crucial module.

The contributions of this work are summarized into three folds: 
\begin{itemize}
\item We propose a dual-perspective semantic-aware representation blending (DSRB) framework, which blends category-specific semantic representation across different images from both instance and prototype perspectives to transfer information of known labels to complement corresponding unknown labels. It does not rely on sufficient annotations and thus performs consistently well on all known label proportions.

\item We design an instance-perspective representation blending (IPRB) module to blend the representation of the known labels in an image to the representation of the corresponding unknown labels in another image to complement these unknown labels, and a prototype-perspective representation blending (PPRB) module to learn more robust representation prototypes and blends the representation of unknown labels with the prototypes of corresponding labels, in a location-sensitive manner, to complement these unknown labels.

\item We conduct extensive and fair experiments on variant benchmark datasets (e.g., Microsoft COCO \cite{Lin2014COCO}, Visual Genome \cite{Krishna2017VG} and Pascal VOC \cite{Everingham2010Pascal}) to demonstrate the effectiveness of the proposed DSRB. For in-depth understanding, we also perform ablative studies to analyze the actual contribution of each crucial module. Codes are available at \textbf{\url{https://github.com/HCPLab-SYSU/HCP-MLR-PL}}.
\end{itemize}

The rest of this paper is organized as follows. We review the works that are related to multi-label image recognition and blending regularization in Section \ref{sec:related}. We seek quantitative insights on the image blending in Section \ref{sec:rethinking} and then introduce the proposed DSRB in Section \ref{sec:method}. We present the experiments and evaluations in Section \ref{sec:experiments} and finally make a conclusion in Section \ref{sec:conclusion}.

\section{Related Work} \label{sec:related}

\noindent{\textbf{MLR with Complete/Partial Labels. }} 
Multi-label image recognition \cite{Wu2020AdaHGNN, Ridnik2021ASL, Gao2021MCAR, Chen2022KGGR} recently receives increasing attention in the computer vision community, because it benefits wide downstream applications, e.g., recommendation systems \cite{Carrillo2013Multi, Zheng2014Context}, scene recognition \cite{Chen2019RoadScene, zhang2020relational, LiuWL15tcyb}, content-based image retrieval \cite{Li2010Technique, Zhang2021Instance}, human attribute recognition \cite{Guo2019Visual, Zhu2017Multi,Chen2021Cross}, etc. Although exploring object localization technology \cite{wei2016hcp} or visual attention mechanism \cite{wang2017multi,chen2018recurrent} to find out potential object regions and enhance feature representation learning, previous works still underperform on most MLR benchmark datasets and even suffer from dramatic performance degradation on the dataset which has many categories, e.g., Visual Genome \cite{Krishna2017VG}. Considering the guidance of semantics to visual representation learning \cite{chen2021hsva}, current works further introduce semantics to enhance category-specific representation learning \cite{Chen2019SSGRL, Wu2020AdaHGNN}, e.g., Semantic Decoupling (SD) module \cite{Chen2019SSGRL, Wu2020AdaHGNN}, Semantic Attention Module (SAM) \cite{Ye2020ADD-GCN} and Class Activation Maps (CAM) \cite{Gao2021MCAR}. On the other hand, label correlations exist commonly among different categories and these correlations are also helpful for multi-label recognition. Thus, recent works resort to graph neural networks \cite{AbadalJGLA22csur,ChenCHWLL20aaai} to explicitly model these correlations to learn contextualized feature representation to facilitate multi-label recognition \cite{Chen2019ML-GCN, Chen2019SSGRL, Wu2020AdaHGNN, Ye2020ADD-GCN, Chen2022KGGR}.

Despite achieving impressive progress, current multi-label image recognition algorithms heavily depends on large-scale datasets with complete and clean annotations per image, making collecting large-scale datasets extremely time-consuming and labor-intensive. To reduce the annotation cost, the current efforts \cite{Durand2019CVPR, Huynh2020CVPR, Chen2022SST, Pu2022SARB} are dedicated to the MLR-PL task, in which merely a few labels are known while the others are unknown for each image (see Figure \ref{fig:task}). Earlier works \cite{Sun2017ICCV, Joulin2016ECCV} formulate MLR as multiple binary classifications, and simply ignore missing labels or treat missing labels as negative. Then, they adopt traditional multi-label recognition algorithms for this task, which leads to poor performance since such an adaptation loses some data or even incurs noisy labels. Inversely, current works tend to generate pseudo labels to complement the unknown labels. For example, Durand et al. \cite{Durand2019CVPR} pre-train image classification models with the given annotations and generate pseudo labels for the unknown labels based on these pre-trained models. Then, they use both the given and updated labels to re-train the models. Huynh et al. \cite{Huynh2020CVPR} propose to train image-level similarity models to generate pseudo labels and progressively re-train the model similarly. However, these algorithms rely on sufficient multi-label annotations for model training, leading to poor performance when the known label proportions decrease to a low level.

Different from these MLR-PL algorithms, our DSRB proposes to blend multi-granularity category-specific semantic representation across different images to complement the unknown labels. Specifically, it consists of two complementary modules in which the first module blends the representation from instance-perspective to generate diverse blended samples while the second module blends semantic representation from prototype-perspective to generate stable blended samples. In this way, the proposed algorithm performs consistently well on the all known label proportions, even when the known label proportion is merely 10\%.

\noindent{\textbf{Blending Regularization. }}
As a blending regularization algorithm, Mixup \cite{Zhang2017Mixup} proposes to perform pixel-wise blending between two images to generate more diverse samples to regularize training. Utilizing this blending regularization, many single-label image recognition models achieve impressive performance. Following the Mixup, Cutmix \cite{Yun2019Cutmix} further proposes an effective regional dropout regularization strategy in which randomly cut one attentional region from an image and paste it into another image to generate a new sample. Furthermore, PuzzleMix \cite{Kim2020Puzzlemix} proposes to explicitly utilize the saliency information and the underlying statistics of the natural examples. Despite the acknowledged progress on single-label image recognition, most blending regularization algorithms are hard to adapt to multi-label image scenarios without extra object location information, e.g., Cutmix, PuzzleMix, etc. Although Mixup can simply blend two multi-label images in image pixel space or global feature map, it easily confuses the contexts in the input images and confuses the representation among different categories, resulting in poor performance.

Different from current blending regularization algorithms, our DSRB proposes to blend multi-granularity category-specific semantic representation across different images to address the above problems. To be specific, our DSRB first learns semantic representation for each category and then blends representation of the same category across different images to transfer information of known labels to complement unknown labels.  

\section{Rethinking Image Blending for MLR} \label{sec:rethinking}
As an effective data augmentation, recently image blending \cite{Zhang2017Mixup, Yun2019Cutmix, gao2022dynamic_mixup, Chang2020Mixup-cam} is widely used to regularize the training process by generating variant training samples in computer vision tasks (e.g., image classification). Despite rapid progress in the single-label image recognition task, image blending still underperforms in the multi-label image recognition task. That is because, as above discussed in Section \ref{sec:intro}, blending two multi-label images via naive mixup operation \cite{Zhang2017Mixup} may lead to semantic and context confusion. To be specific, the root causes of this problem are as follows: 1) Compared with the single-label counterparts, the multi-label images naturally contain multiple semantic objects of diverse semantic categories that have variant sizes and appearances and scatter over the whole image. Thus, simply blending two multi-label images easily mix two objects with totally different semantics together, which harms the semantic representation learning of the model. 2) As explored in recent leading algorithms, the context dependencies can provide extra guidance to facilitate multi-label recognition. However, blending two images, especially if they belong to different scenes, may destroy these dependencies and thus provide confusing context information for recognizing original semantic objects. 

To avoid the above-mentioned semantic and context confusion, it is may effective that blend two multi-label images in the category-specific semantic representation space rather than in the image pixel space. We argue that category-specific semantic representation blending not only avoids mixing semantics from different categories but also retains original context information for recognizing objects. It is worth noting that the purpose of the category-specific semantic representation blending, different from Mixup, is to complement the unknown labels in the MLR-PL task rather than regularize the training process. 

\begin{figure*}[!t]
   \centering
   \includegraphics[width=0.95\linewidth]{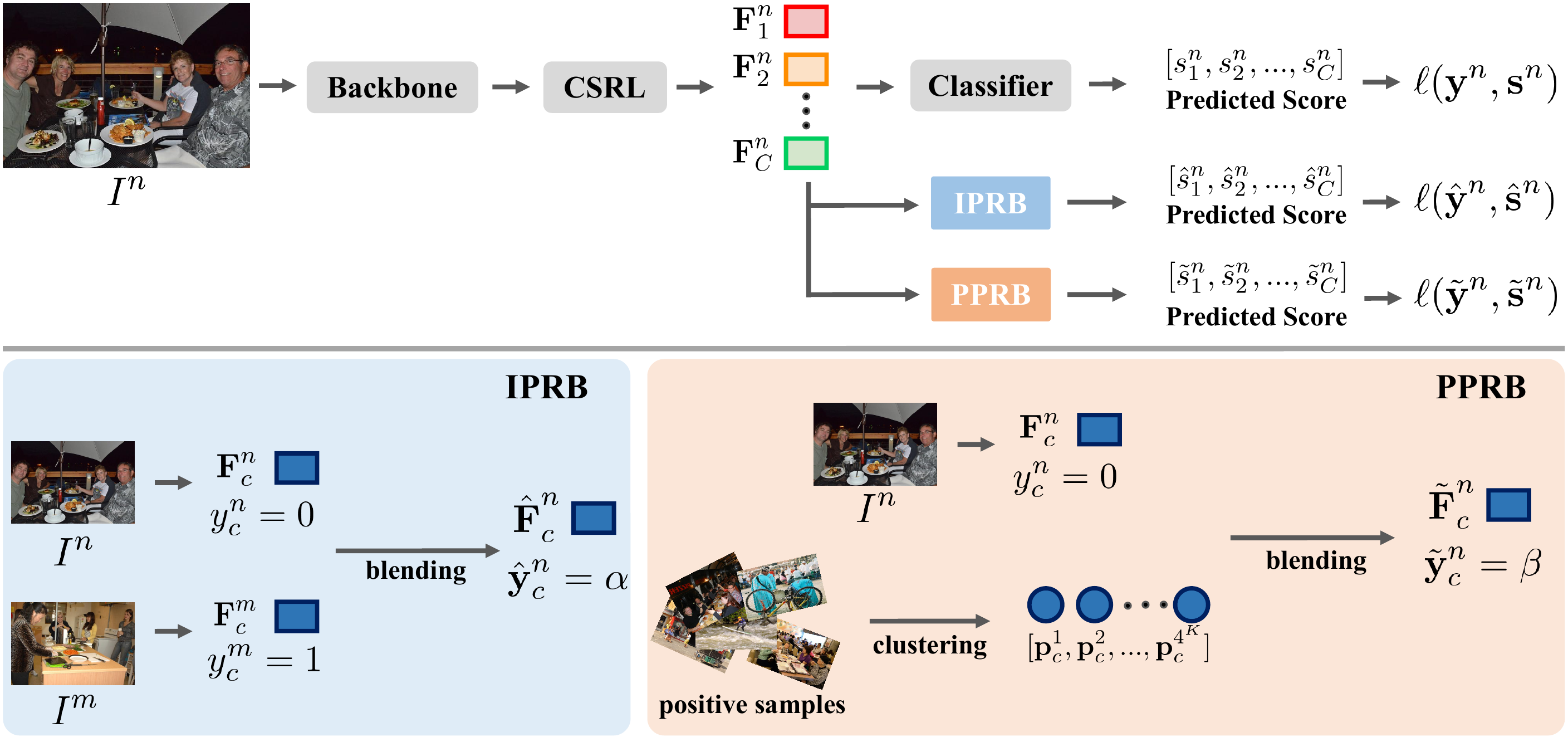}
   \caption{An overall illustration of the proposed dual-perspective semantic-aware representation blending (DSRB). The upper part is the overall pipeline that consists of the IPRB and PPRB modules that perform instance-perspective and prototype-perspective representation blending to complement unknown labels. The lower part is the detailed implementations of the IPRB and PPRB modules. The IPRB module blends the semantic representations of the known labels in an image $I^m$ to the representations of the corresponding unknown labels in another image $I^n$ to complement these unknown labels. The PPRB module learns more stable representation prototypes for each category and blends the representation of unknown labels with the prototypes of corresponding labels to complement these unknown labels. Finally, the known labels and generated pseudo labels are used to supervise the training of the multi-label recognition model.}
   \label{fig:framework}
\end{figure*}

\section{Dual-perspective Semantic-aware Representation Blending} \label{sec:method}

\subsection{Overview}
In this section, we introduce the proposed DSRB that blends category-specific semantic representation across different images to transfer information of known labels to help complement unknown labels. The proposed DSRB consists of two complementary representation blending modules, i.e., the instance-perspective representation blending (IPRB) and prototype-perspective representation blending (PPRB), that perform instance-perspective and prototype-perspective representation blending to transfer information of known labels across different images. Specifically, the IPRB module blends the semantic representations of known labels in one image to the semantic presentations of the unknown labels in another image to complement these unknown labels. Meanwhile, the PPRB module learns more robust representation prototypes for each category and generates more stable blended representations based on potential object presence area to complement unknown labels. Finally, both the ground truth and generated pseudo labels are used to train the multi-label model to facilitate the MLR-PL task. Figure \ref{fig:framework} illustrates an overall pipeline of the proposed DSRB.

\noindent\textbf{Notation. } Follow prior MLR-PL work \cite{Durand2019CVPR}, we denote by $C$ the number of categories and $N$ the number of training examples. We denote the training dataset as $\mathcal{D}=\{(I^1, \textbf{y}^1), ..., (I^N, \textbf{y}^N)\}$, where $I^n$ is the $n$-th image and $\textbf{y}^n=\{y^n_1, y^n_2, \cdots, y^n_C\}\in \{-1, 0, 1\}^C$ is the corresponding label vector. Specifically, $y^n_c$ is assigned to 1 if label $c$ exists in the $n$-th image, assigned to -1 if it does not exist, and assigned to 0 if it is unknown.

\subsection{Category-specific Representation Learning}
The category-specific representation learning (CSRL) module is introduced to learn category-specific semantic feature maps by incorporating semantic information of each category. Given a training image $I^n$, we first utilize a backbone network to extract the global feature maps $\textbf{f}^n$, then utilize the CSRL module that pays more attention to the corresponding object regions to extract category-specific semantic feature maps and perform pooling operation to obtain category-specific semantic representation vectors. Formulated as 
\begin{gather}
    [\textbf{F}^n_1, \textbf{F}^n_2, \cdots, \textbf{F}^n_C] = \phi_{csrl}(\textbf{f}^n), \\
    \textbf{f}^n_c = \sigma(\textbf{F}^n_c) \quad for \quad c \in [1, C],
\end{gather}
where $\phi_{csrl}$ is the CSRL module, $\sigma$ is the pooling operation, $\textbf{F}^n_c$ is the category-specific semantic feature maps of the category $c$ in $n$-th image and $\textbf{f}^n_c$ is the corresponding category-specific representation vector. 

Here, we utilize current algorithms to implement the CSRL module, i.e., semantic decoupling \cite{Chen2019SSGRL} and semantic attention \cite{Ye2020ADD-GCN}. The semantic decoupling \cite{Chen2019SSGRL} introduces a semantic embedding vector for each category and then incorporates this vector to guide learning feature maps for each category that assigns higher attention values to potential object presence areas. The semantic attention \cite{Ye2020ADD-GCN} generates the category-specific activation maps by using class activation mapping \cite{Zhou2016CAM}, then utilize these maps to convert the transformed feature map into the content-aware feature maps for each category.

Compared with the single-label counterpart, the multi-label images naturally contain multiple semantic objects scattering over the whole image and complex background, therefore the semantic representation of each category likely contains the semantics of other categories and information of complex background. Directly performing blending operation on these semantic representations easily generate confusing training samples, thus we need to learn more compact category-specific semantic representation. To achieve this end, we utilize contrastive loss for increasing the similarity between $\textbf{f}^n_c$ and $\textbf{f}^m_c$ if images $n$ and $m$ have the same existing category $c$ and decreasing the similarity otherwise. Thus, it can be formulated as
\begin{equation}
\ell^{n,m}_c=
    \begin{cases}
         1-cosine(\textbf{f}^n_c, \textbf{f}^m_c) \quad & y^n_c=1,y^m_c=1,\\
         1+cosine(\textbf{f}^n_c, \textbf{f}^m_c) \quad & otherwise,
    \end{cases}
\end{equation}
where $cosine(\cdot, \cdot)$ represents a function that computes the cosine similarity between the input. The final contrastive loss can be formulated as 
\begin{equation}
\mathcal{L}_{cst}=\sum^{N}_{n=1} \sum^{N}_{m=1} \sum^{C}_{c=1} \ell^{n,m}_c.
\end{equation}

Once obtain the category-specific representation vector, we follow previous work \cite{Chen2019SSGRL,Chen2022KGGR, Ye2020ADD-GCN, Chen2021Cross} to use a gated neural network and a linear classifier followed by a sigmoid function to compute the probability score vectors
\begin{equation}
  [s^n_1, s^n_2, \cdots, s^n_C]=\phi([\textbf{f}^n_1, \textbf{f}^n_2, \cdots, \textbf{f}^n_C]).
\end{equation}

\subsection{Instance-perspective Representation Blending}
Despite achieving impressive progress, current MLR-PL algorithms suffer from obvious performance drop if decreasing the known label proportion to a small level because these algorithms depend on sufficient annotations to train image classification or similarity model to generate pseudo labels. To avoid such constraints, the IPRB module is designed to blend the information of label $c$ in image $I^m$ to image $I^n$ to complement unknown labels which still works when known labels are few, thus the model performs consistently well on all known label proportion settings. To achieve this end, we blend the semantic feature maps that belong to the same category and from different images to transfer the known labels of one image to the unknown labels of the other image.

Formally, given two training images $I^n$ and $I^m$, whose learned semantic feature maps are $\textbf{F}^{n}=[\textbf{F}^{n}_1, \textbf{F}^{n}_2, \cdots, \textbf{F}^{n}_C]$ and $\textbf{F}^{m}=[\textbf{F}^{m}_1, \textbf{F}^{m}_2, \cdots, \textbf{F}^{m}_C]$, and label vectors are $\textbf{y}^n=\{y^n_1, y^n_2, \cdots, y^n_C\}$ and $\textbf{y}^m=\{y^m_1, y^m_2, \cdots, y^m_C\}$, we blend the semantic feature maps and label for each category based on label vectors. For category $c$, the blending process can be formulated as 
\begin{equation}
 \hat{\textbf{F}}^{n}_c=
  \begin{cases}     
   \alpha \textbf{F}^{n}_c + (1-\alpha) \textbf{F}^{m}_c \quad &  y^n_c=0, y^m_c=1, \\
   \textbf{F}^{n}_c \quad & otherwise,
  \end{cases}
\end{equation}

\begin{equation}
 \hat{\textbf{y}}^{n}_c=
  \begin{cases}
   1 - \alpha \quad & y^n_c=0, y^m_c=1, \\
   y^{n}_c \quad & otherwise,\\
  \end{cases}
\end{equation}
where $\alpha$ is the learnable parameter and its initial value is set to 0.5. We repeat the above blending process for all categories, and reformulate them as matrix operations for efficient computing 
\begin{equation}
 {\hat{\textbf{F}}^{n}} = A \textbf{F}^{n} + (1-A) \textbf{F}^{m},
 \end{equation}
\begin{equation}
 {\hat{\textbf{y}}^{n}} = A \textbf{y}^n + (1-A) \textbf{y}^m,
\end{equation}
where $A=[\alpha_1, \alpha_2, \cdots, \alpha_C]$ is a parameter vector; $\hat{\textbf{F}}^{n}=[\hat{\textbf{F}}^{n}_1, \hat{\textbf{F}}^{n}_2, \cdots, \hat{\textbf{F}}^{n}_C]$ and $\hat{\textbf{y}}^{n}=[\hat{y}^{n}_1, \hat{y}^{n}_2, \cdots, \hat{y}^{n}_C]$ are the blended semantic representation feature maps and label matrix. Then, we operate pooling to these feature maps to obtain blended semantic representation vectors, and then feed it into a gated graph neural network and linear classifier followed by sigmoid function to compute the probability score vector $\hat{\textbf{s}}^{n}$.

\subsection{Prototype-perspective Representation Blending}
\begin{figure}[!h]
   \centering
   \includegraphics[width=0.95\linewidth]{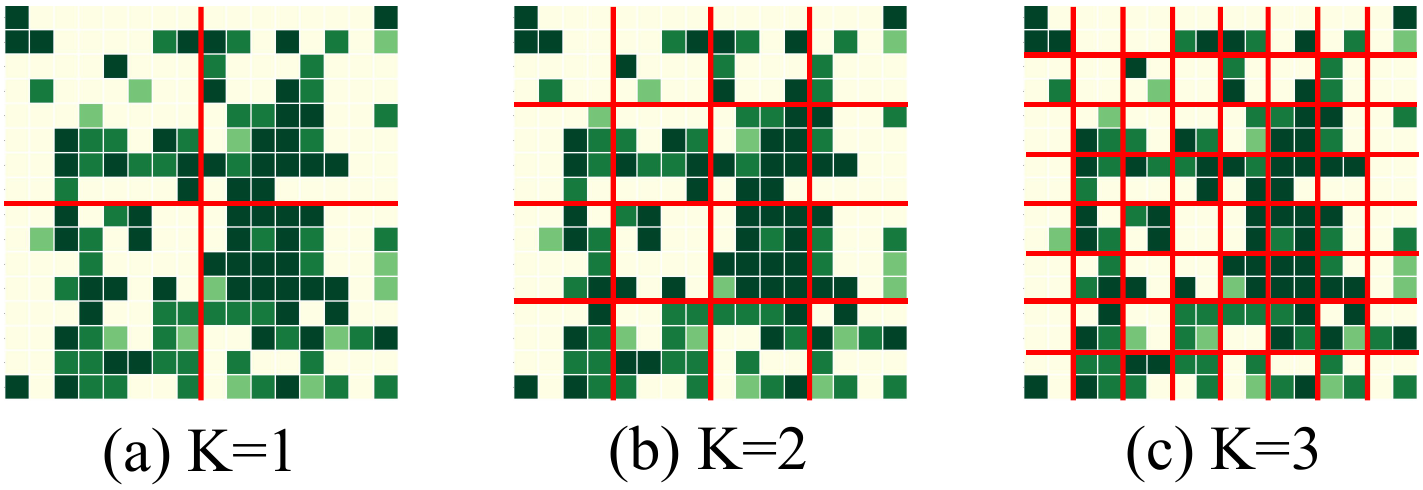}
   \caption{Several examples of dividing feature maps over $x$ and $y$ axis with different values of $K$. As shown, with the increase of the value of $K$, the localization information of objects is more fine-grained while the computation cost increase obviously.}
   \label{fig:loc-example}
\end{figure}

Considering multi-label images inherently possesses multiple semantic objects scattering over the whole image, simply blending two samples may generate confusing semantic information which may disturb slightly the training process. To deal with this issue, we further design the PPRB module that learns more robust representation prototypes for each category and generates more stable blended semantic feature maps based on potential object regions. For each category $c$, the PPRB module first select all the images that have the known label $c$, and then extract the corresponding feature maps of this category, resulting in the feature maps set $[\textbf{F}^{1}_c, \textbf{F}^{2}_c, \cdots, \textbf{F}^{N_c}_c]$ where $N_c$ is the number of positive samples of category $c$. As shown in Figure \ref{fig:loc-example}, the PPRB module divide these semantic feature maps into $4^K$ subsets based on maximum of feature maps over $x$ and $y$ axis, obtains $4^K$ representation prototypes by averaging these subsets on channel, i.e., $\textbf{P}_{c}=[\textbf{p}^1_c, \textbf{p}^2_c, ..., \textbf{p}^{4^K}_c]$.

Given an input image $I^n$ whose learned semantic representation maps $\textbf{F}^{n}=[\textbf{F}^{n}_1, \textbf{F}^{n}_2, \cdots, \textbf{F}^{n}_C]$ and corresponding label vectors $\textbf{y}^n=\{y^n_1, y^n_2, \cdots, y^n_C\}$, we randomly select a category $c$ whose label is unknown, then adaptively select a representation prototype from $\textbf{P}_{c}$ based on potential object areas and blend it with the semantic feature maps of category $c$, formulated as 
\begin{equation}
 \tilde{\textbf{F}}^{n}_c=
  \begin{cases}
   \beta \textbf{F}^{n}_c + (1-\beta) \textbf{P}^{k}_c \quad & c=random(\{c|y^n_c=0\}) \\
   \textbf{F}^{n}_c \quad & otherwise,\\
  \end{cases}
\end{equation}

\begin{equation}
 \tilde{\textbf{y}}^{n}_c=
  \begin{cases}
   1 - \beta & c=random(\{c|y^n_c=0\})\\
   y^{n}_c  & otherwise,\\
  \end{cases}
\end{equation}
where $\beta$ is also a learnable parameter, and it is initialized as 0.5; $random(\cdot)$ represents a random sampling function which means we randomly choose one unknown category to blend semantic representation per image; $k$ is randomly sampled in $[1,...,4^K]$ which obeys uniform distribution and avoid the potential object area of prototype is same with input image. We repeat the above blending process for all categories, and reformulate them as matrix operations for efficient computing: 
\begin{equation}
 {\tilde{\textbf{F}}^{n}} = B \textbf{F}^{n} + (1-B) \textbf{P}^{k},
 \end{equation}
\begin{equation}
 {\tilde{\textbf{y}}^{n}} = B \textbf{y}^n + (1-B),
\end{equation}
where $B=[\beta_1, \beta_2, \cdots, \beta_C]$ is a parameter vector; $\textbf{P}^{k}=[\textbf{p}^{k_1}_1, \textbf{p}^{k_2}_2, \cdots, \textbf{p}^{k_C}_C]$ are representation prototypes; $\tilde{\textbf{F}}^{n}=[\tilde{\textbf{F}}^{n}_1, \tilde{\textbf{F}}^{n}_2, \cdots, \tilde{\textbf{F}}^{n}_C]$ and $\tilde{\textbf{y}}^{n}=[\tilde{\textbf{y}}^{n}_1, \tilde{\textbf{y}}^{n}_2, \cdots, \tilde{\textbf{y}}^{n}_C]$ are the blended semantic representation feature maps and label matrix.  we operate pooling to these feature maps to obtain blended semantic representation vectors, and then feed it into a gated graph neural network and linear classifier followed by sigmoid function to compute the probability score vector $\tilde{\textbf{s}}^{n}$.

\begin{figure}[!h]
   \centering
   \includegraphics[width=0.95\linewidth]{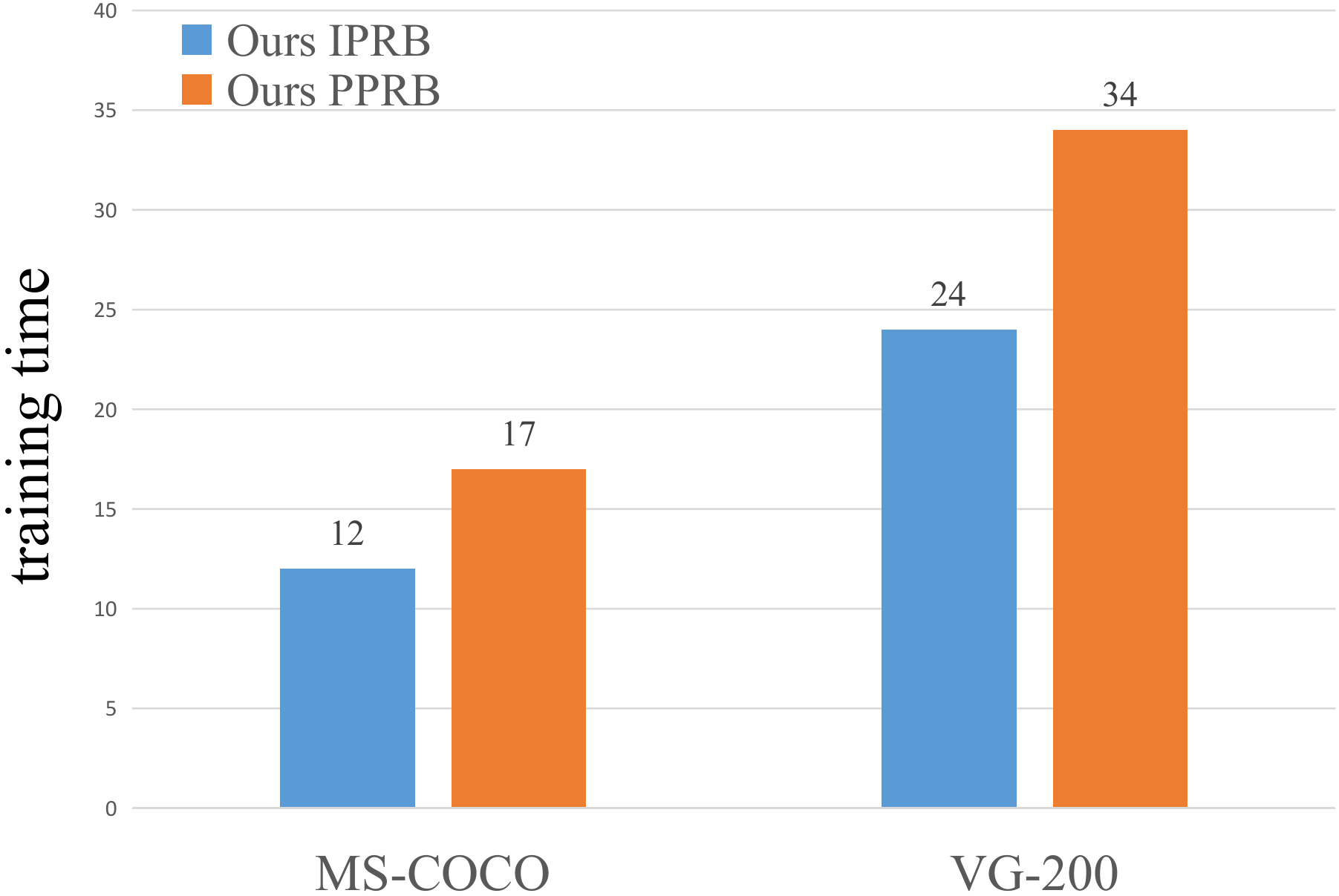}
   \caption{The training time (hours) of the IPRB module (denoted as ``Ours IPRB") and the PPRB module (denoted as ``Ours PPRB") on the 50\% known labels settings on MS-COCO (left) and VG-200 (right) datasets.}
   \label{fig:training-time}
\end{figure}

\subsection{Discussions on IPRB and PPRB}
As discussed above, the IPRB and PPRB are two complementary modules that transfer information of known labels to help complement unknown labels. To be specific, we propose to blend the semantic of  corresponding category between two different images in the IPRB module and propose to blend the semantics of corresponding category between images and prototypes in the PPRB module. Compared with the PPRB, the IPRB can blends the corresponding semantic of two different images to generate more diverse training samples. However, it may generates confusing samples when the known labels are few because IPRB does not limit the object position of the corresponding category in these two different images which may confuse the semantic of part categories. However, such a limitation dramatically reduces the available blending images and results in poor performance. Therefore, we do not introduce the limitation in the IPRB. Besides, although the PPRB brings more performance improvement than the IPRB, the PPRB brings more computational cost and training time, as shown in Figure \ref{fig:training-time}.

In summary, the IPRB module can generates diverse training samples by blending the semantics among different images, but may generates confusing training samples when the known labels are few. Meanwhile, the PPRB module can generate stable training samples by blending the semantics between images and prototypes, but its computational cost and training time are higher than the IPRB.

\subsection{Optimization}
Following previous works, we utilize the partial binary cross entropy loss as the objective function for supervising the network. In particular, given the predicted probability score vector $\textbf{s}^n=\{s^n_1, s^n_2, \cdots s^n_C\}$ and the ground truth of known labels, the objective function can be defined as 
\begin{equation}
\begin{aligned}
\ell(\textbf{y}^n, \textbf{s}^n)=&\frac{1}{\sum_{c=1}^C|y^n_c|}\sum_{c=1}^C[\textbf{1}(y^n_c=1)\log(s^n_c) \\
&+\textbf{1}(y^n_c=-1)\log(1-s^n_c)],
\end{aligned}
\end{equation}
where $\textbf{1}[\cdot]$ is an indicator function whose value is 1 if the argument is positive and is 0 otherwise.

Similarly, we adopt the partial binary cross entropy loss as the objective function for supervising the IPRB module and PPRB module, i.e., $\ell(\hat{\textbf{y}}^{n}, \hat{\textbf{s}}^{n})$ and $\ell(\tilde{\textbf{y}}^{n}, \tilde{\textbf{s}}^{n})$. Therefore, the final classification loss is defined as summing the three losses over all samples, formulated as
\begin{equation}
 \begin{aligned}
  \mathcal{L}_{cls} &= \sum^{N}_{n=1}{ [ \ell(\textbf{y}^n, \textbf{s}^n) + \ell(\hat{\textbf{y}}^{n}, \hat{\textbf{s}}^{n}) + \ell(\tilde{\textbf{y}}^{n}, \tilde{\textbf{s}}^{n}) ] }.
 \end{aligned}
\label{eq:cls-loss}
\end{equation}

Finally, we sum over the classification and contrastive losses of all samples to obtain the final loss, formulated as
\begin{equation}
\mathcal{L}=\mathcal{L}_{cls} + \lambda \mathcal{L}_{cst}.
\label{eq:total-loss}
\end{equation}
Here, $\lambda$ is a balance parameter that ensures the contrastive loss $\mathcal{L}_{cst}$ has a comparable magnitude with the classification loss $\mathcal{L}_{cls}$. Since $\mathcal{L}_{cst}$ is much larger than $\mathcal{L}_{cls}$, we set $\lambda$ to 0.05 in the experiments.

\section{Experiments} \label{sec:experiments}

\subsection{Experimental Setting}

\subsubsection{Implementation Details} 
For fair comparison, we follow previous works \cite{Chen2019SSGRL, Durand2019CVPR} to adopt the ResNet-101 \cite{He2016ResNet} as the backbone to extract global feature maps $\textbf{f}^n$. We initialize its parameters with those pre-trained on the ImageNet \cite{Deng2009Imagenet} dataset while initializing the parameters of all newly-added layers randomly. We fix the parameters of the previous 91 layers of ResNet-101, and train the other layers in an end-to-end manner. During training, we use the Adam algorithm \cite{Kingma2015Adam} with a batch size of 32, momentums of 0.999 and 0.9, and a weight decay of $5 \times 10^{-4}$. We set the initial learning rate as $10^{-5}$ and divide it by 10 after every 10 epochs. It is trained with 20 epochs in total. For data augmentation, the input image is resized to 512$\times$512, and we randomly choose a number from \{512, 448, 384, 320, 256\} as the width and height to crop patch. Finally, the cropped patch is further resized to 448$\times$448. Besides, random horizontal flipping is also used. To stabilize the training process, we start to use the IPRB and PPRB modules at epoch 5, and re-compute prototypes of each category for every 5 epochs. During inference, the IPRB and PPRB modules are removed, and the image is resized to 448$\times$448 for evaluation. To facilitate the practical implementation, we set $K$ to 1 in the experiments. 

\subsubsection{Dataset} We conduct experiments on the MS-COCO \cite{Lin2014COCO}, Visual Genome \cite{Krishna2017VG}, and Pascal VOC 2007 \cite{Everingham2010Pascal} datasets for fair comparison. MS-COCO covers 80 daily-lift categories, which contains 82,801 images as the training set and 40,504 images as the validation set. Pascal VOC 2007 contains 9,963 images from 20 object categories, and we follow previous works to use the trainval set for training and the test set for evaluation. Visual Genome contains 108,249 images from 80,138 categories, and most categories have very few samples. In this work, we select the 200 most frequent categories to obtain a VG-200 subset. Moreover, since there is no train/val split, we randomly select 10,000 images as the test set and the rest 98,249 images are used as the training set. The train/test set will be released for further research.

Since all the datasets have complete labels, we follow the setting of previous works \cite{Durand2019CVPR, Huynh2020CVPR} to randomly drop a certain proportion of positive and negative labels to create partially annotated datasets. In this work, the proportions of dropped labels vary from 90\% to 10\%, resulting in known labels proportion of 10\% to 90\%. 

\subsubsection{Evaluation Metric} For a fair comparison, we follow previous works \cite{Durand2019CVPR, Huynh2020CVPR} to adopt the mean average precision (mAP) over all categories for evaluation under different proportions of known labels. And we also compute average mAP over all proportions for a more comprehensive evaluation. Moreover, we follow most previous MLR works \cite{Chen2019SSGRL} to adopt the overall and per-class F1-measure (i.e., OF1 and CF1) for a more comprehensive evaluation. Formally, the OF1 and CF1 can be computed by
\begin{gather}
 OP=\frac{\sum_{i}{N^c_i}}{\sum_{i}{N^p_i}}, CP=\frac{1}{C} \sum_{i}{ \frac{N^c_i}{N^p_i} } \\
 OR=\frac{\sum_{i}{N^c_i}}{\sum_{i}{N^g_i}}, CR=\frac{1}{C} \sum_{i}{ \frac{N^c_i}{N^g_i} } \\
 OF1=\frac{2 \times OP \times OR }{OP+OR},  CF1=\frac{2 \times CP \times CR }{CP+CR}
\end{gather}
where $N^c_i$ is the number of images that are correctly predicted for the $i$-th label, $N^p_i$ is the number of predicted images for the $i$-th label, $N^g_i$ is the number of ground truth images for the $i$-th label. We also average the OF1, CF1 on all known label proportion settings.

\begin{table}[!h]
  \centering
  \begin{tabular}{@{\hspace{0.5ex}}c@{\hspace{0.5ex}}|@{\hspace{0.5ex}}c@{\hspace{0.5ex}}|cc}
  \hline
  \centering Datasets & Methods & Avg. OF1 & Avg. CF1 \\
  \hline
  \hline
  \centering \multirow{8}*{MS-COCO} & SSGRL & 73.9 & 68.1 \\
  \centering ~ & GCN-ML & 73.1 & 68.4 \\
  \centering ~ & KGGR & 73.7 & 69.7 \\
  \centering ~ & P-GCN & - & - \\
  \centering ~ & ASL & 46.7 & 47.9 \\
  \centering ~ & CL & 61.9 & 48.3 \\
  \centering ~ & Partial BCE & 74.0 & 68.8 \\
  \centering ~ & DSRB & \textbf{76.8} & \textbf{72.7} \\
  \hline
  \hline
  \centering \multirow{8}*{VG-200} & SSGRL & 37.8 & 26.1 \\
  \centering ~ & GCN-ML & 38.7 & 25.6 \\
  \centering ~ & KGGR & 41.2 & 33.6 \\
  \centering ~ & P-GCN & - & - \\
  \centering ~ & ASL & 24.9 & 23.3 \\
  \centering ~ & CL & 23.6 & 10.9 \\
  \centering ~ & Partial BCE & 36.1 & 25.7 \\
  \centering ~ & DSRB & \textbf{45.1} & \textbf{37.7} \\
  \hline
  \hline
  \centering \multirow{8}*{Pascal VOC 2007} & SSGRL & 87.7 & 84.5 \\
  \centering ~ & GCN-ML & 87.3 & 84.6 \\
  \centering ~ & KGGR & 86.5 & 84.7 \\
  \centering ~ & P-GCN & - & - \\
  \centering ~ & ASL & 41.0 & 40.9 \\
  \centering ~ & CL & 83.8 & 75.4 \\
  \centering ~ & Partial BCE & 87.9 & 84.8 \\
  \centering ~ & DSRB & \textbf{88.3} & \textbf{86.0} \\
  \hline
  \end{tabular}
  \vspace{10pt}
  \caption{Average mAP, OF1 and CF1 of the proposed DSRB and current state-of-the-art competitors for multi-label recognition with partial labels on the MS-COCO, VG-200 and Pascal VOC 2007 datasets. The best results are highlighted in bold. ``-" denotes the corresponding result in not provided.}
  \label{tab:average-results}
\end{table}

\begin{table*}
  \centering
  \small
  \begin{tabular}{c|c|ccccccccc|c}
  \hline
  \centering Datasets & Methods & 10\% & 20\% & 30\% & 40\% & 50\% & 60\% & 70\% & 80\% & 90\% & Ave. mAP \\
  \hline
  \hline
  \centering \multirow{8}*{MS-COCO} & SSGRL & 62.5 & 70.5 & 73.2 & 74.5 & 76.3 & 76.5 & 77.1 & 77.9 & 78.4 & 74.1 \\
  \centering ~ & GCN-ML & 63.8 & 70.9 & 72.8 & 74.0 & 76.7 & 77.1 & 77.3 & 78.3 & 78.6 & 74.4 \\
  \centering ~ & KGGR & 66.6 & 71.4 & 73.8 & 76.7 & 77.5 & 77.9 & 78.4 & 78.7 & 79.1 & 75.6 \\
  \centering ~ & P-GCN & 67.5 & 71.6 & 73.8 & 75.5 & 77.4 & 78.4 & 79.5 & \textbf{80.7} & \textbf{81.5} & 76.2 \\
  \centering ~ & ASL & 69.7 & 74.0 & 75.1 & 76.8 & 77.5 & 78.1 & 78.7 & 79.1 & 79.7 & 76.5 \\
  \centering ~ & CL & 26.7 & 31.8 & 51.5 & 65.4 & 70.0 & 71.9 & 74.0 & 77.4 & 78.0 & 60.7 \\
  \centering ~ & Partial BCE & 61.6 & 70.5 & 74.1 & 76.3 & 77.2 & 77.7 & 78.2 & 78.4 & 78.5 & 74.7 \\
  \centering ~ & DSRB & \textbf{72.5} & \textbf{76.0} & \textbf{77.6} & \textbf{78.7} & \textbf{79.6} & \textbf{79.8} & \textbf{80.0} & 80.5 & 80.8 & \textbf{78.4} \\
  \hline
  \hline
  \centering \multirow{8}*{VG-200} & SSGRL & 34.6 & 37.3 & 39.2 & 40.1 & 40.4 & 41.0 & 41.3 & 41.6 & 42.1 & 39.7 \\
  \centering ~ & GCN-ML & 32.0 & 37.8 & 38.8 & 39.1 & 39.6 & 40.0 & 41.9 & 42.3 & 42.5 & 39.3 \\
  \centering ~ & KGGR & 36.0 & 40.0 & 41.2 & 41.5 & 42.0 & 42.5 & 43.3 & 43.6 & 43.8 & 41.5 \\
  \centering ~ & P-GCN & - & - & - & - & - & - & - & - & - & - \\
  \centering ~ & ASL & 38.4 & 39.5 & 40.6 & 40.6 & 41.1 & 41.4 & 41.4 & 41.6 & 41.6 & 40.7 \\
  \centering ~ & CL & 12.1 & 19.1 & 25.1 & 26.7 & 30.0 & 31.7 & 35.3 & 36.8 & 38.5 & 28.4 \\
  \centering ~ & Partial BCE & 27.4 & 38.1 & 40.2 & 40.9 & 41.5 & 42.1 & 42.4 & 42.7 & 42.7 & 39.8 \\
  \centering ~ & DSRB & \textbf{41.4} & \textbf{44.0} & \textbf{44.8} & \textbf{45.5} & \textbf{46.6} & \textbf{47.5} & \textbf{47.8} & \textbf{48.0} & \textbf{48.2} & \textbf{46.0} \\
  \hline
  \hline
  \centering \multirow{8}*{VOC 2007} & SSGRL & 77.7 & 87.6 & 89.9 & 90.7 & 91.4 & 91.8 & 92.0 & 92.2 & 92.2 & 89.5 \\
  \centering ~ & GCN-ML & 74.5 & 87.4 & 89.7 & 90.7 & 91.0 & 91.3 & 91.5 & 91.8 & 92.0 & 88.9 \\
  \centering ~ & KGGR & 81.3 & 88.1 & 89.9 & 90.4 & 91.2 & 91.3 & 91.5 & 91.6 & 91.8 & 89.7 \\
  \centering ~ & P-GCN & 82.5 & 85.4 & 88.2 & 89.8 & 90.0 & 90.9 & 91.6 & 92.5 & 93.1 & 89.3 \\
  \centering ~ & ASL & 82.9 & 88.6 & 90.0 & 91.2 & 91.7 & 92.2 & 92.4 & 92.5 & 92.6 & 90.5 \\
  \centering ~ & CL & 44.7 & 76.8 & 88.6 & 90.2 & 90.7 & 91.1 & 91.6 & 91.7 & 91.9 & 84.1 \\
  \centering ~ & Partial BCE & 80.7 & 88.4 & 89.9 & 90.7 & 91.2 & 91.8 & 92.3 & 92.4 & 92.5 & 90.0 \\
  \centering ~ & DSRB & \textbf{85.7} & \textbf{89.8} & \textbf{91.8} & \textbf{92.0} & \textbf{92.3} & \textbf{92.7} & \textbf{92.9} & \textbf{93.1} & \textbf{93.2} & \textbf{91.5} \\
  \hline
  \end{tabular}
  \vspace{10pt}
  \caption{Performance of our DSRB and current state-of-the-art competitors for MLR-PL on the MS-COCO, VG-200 and Pascal VOC 2007 datasets. The best results are highlighted in bold. ``-" denotes the corresponding result is not provided.}
  \label{tab:mAP-results}
\end{table*}

\begin{table}[!t]
  \centering
  \begin{tabular}{c|c|cc}
  \hline
  \centering Datasets & Methods & Avg. OF1 & Avg. CF1 \\
  \hline
  \hline
  \centering \multirow{10}*{MS-COCO} & SSGRL & 73.9 & 68.1 \\
  \centering ~ & SSGRL w/ IP-Mixup & 72.5 & 67.8 \\
  \centering ~ & SSGRL w/ FM-Mixup & 73.0 & 67.6 \\
  \centering ~ & Ours IPRB & 76.2 & 71.9 \\
  \centering ~ & Ours IPRB w/ fixed $\alpha$ & 75.3 & 70.8 \\
  \centering ~ & Ours PPRB & 76.6 & 72.4 \\
  \centering ~ & Ours PPRB w/ fixed $\beta$ & 75.9 & 71.7 \\
  \centering ~ & Ours PPRB w/ RV & 75.9 & 71.5 \\
  \centering ~ & Ours w/ SA & 76.8 & 72.6 \\
  \centering ~ & Ours w/ RV-Mixup & 76.5 & 72.2 \\
  \centering ~ & Ours & \textbf{76.8} & \textbf{72.7} \\
  \hline
  \hline
  \centering \multirow{10}*{VG-200} & SSGRL & 37.8 & 26.1 \\
  \centering ~ & SSGRL w/ IP-Mixup & 44.5 & 34.5 \\
  \centering ~ & SSGRL w/ FM-Mixup & 38.7 & 29.2 \\
  \centering ~ & Ours IPRB & 43.4 & 35.3 \\
  \centering ~ & Ours IPRB w/ fixed $\alpha$ & 42.9 & 35.2 \\
  \centering ~ & Ours PPRB & 44.5 & 36.7 \\
  \centering ~ & Ours PPRB w/ fixed $\beta$ & 43.1 & 34.8 \\
  \centering ~ & Ours PPRB w/ RV & 44.1 & 36.5 \\
  \centering ~ & Ours w/ SA & 45.0 & 37.5 \\
  \centering ~ & Ours w/ RV-Mixup & 45.0 & 37.4 \\
  \centering ~ & Ours & \textbf{45.1} & \textbf{37.7} \\
  \hline
  \hline
  \centering \multirow{10}*{Pascal VOC 2007} & SSGRL & 87.7 & 84.5 \\
  \centering ~ & SSGRL w/ IP-Mixup & 84.3 & 80.7 \\
  \centering ~ & SSGRL w/ FM-Mixup & 85.9 & 82.9 \\
  \centering ~ & Ours IPRB & 88.1 & 85.5 \\
  \centering ~ & Ours IPRB w/ fixed $\alpha$ & 87.2 & 83.8 \\
  \centering ~ & Ours PPRB & 88.3 & 85.9 \\
  \centering ~ & Ours PPRB w/ fixed $\beta$ & 87.1 & 83.7 \\
  \centering ~ & Ours PPRB w/ RV & 88.0 & 85.1 \\
  \centering ~ & Ours w/ SA & 88.2 & \textbf{86.2} \\
  \centering ~ & Ours w/ RV-Mixup & 88.3 & 85.9 \\
  \centering ~ & Ours & \textbf{88.4} & 86.0 \\
  \hline
  \end{tabular}
  \vspace{10pt}
  \caption{The average OF1 and CF1 of the baseline SSGRL, SSGRL with mixup on image pixel level (IP-Mixup), SSGRL with mixup on feature map level (FM-Mixup), our DSRB merely using IPRB module (Our IPRB), our DSRB merely using IPRB module with fixed $\alpha$ (Ours IPRB w/ fixed $\alpha$), our DSRB merely using PPRB module (Ours PPRB), our DSRB merely using PPRB module with fixed $\beta$ (Ours PPRB w/ fixed $\beta$), our DSRB merely using PPRB module with category-specific representation vector level blending (Ours PPRB w/ RV), our DSRB with mixup on category-specific representation vector level (Ours w/ RV-Mixup), our DSRB using SA instead of SD (Ours w/ SA) and our DSRB (Ours) on the MS-COCO, VG-200 and Pascal VOC 2007 datasets. The best results are highlighted in bold.}
  \label{tab:average-ablation-results}
\end{table}

\begin{table*}
  \centering
  \small
  \begin{tabular}{c|c|ccccccccc|c}
  \hline
  \centering Datasets & Methods & 10\% & 20\% & 30\% & 40\% & 50\% & 60\% & 70\% & 80\% & 90\% & Ave. mAP \\
  \hline
  \hline
  \centering \multirow{10}*{MS-COCO} & SSGRL & 62.5 & 70.5 & 73.2 & 74.5 & 76.3 & 76.5 & 77.1 & 77.9 & 78.4 & 74.1 \\
  \centering ~ & SSGRL w/ IP-Mixup & 61.1 & 70.0 & 73.0 & 74.7 & 77.0 & 76.9 & 77.7 & 78.4 & 78.8 & 74.2 \\
  \centering ~ & SSGRL w/ FM-Mixup & 61.0 & 71.1 & 73.4 & 75.1 & 76.7 & 77.2 & 78.2 & 78.4 & 78.8 & 74.4 \\
  \centering ~ & Ours IPRB & 71.7 & 75.1 & 77.2 & 78.2 & 78.9 & 79.2 & 79.4 & 79.9 & 80.3 & 77.8 \\
  \centering ~ & Ours IPRB w/ fixed $\alpha$ & 70.6 & 74.1 & 76.5 & 77.5 & 78.4 & 78.4 & 78.7 & 79.1 & 79.4 & 77.0 \\
  \centering ~ & Ours PPRB & 72.0 & 75.3 & 77.1 & 78.5 & 79.3 & 79.6 & 79.7 & 79.9 & 80.4 & 78.0 \\
  \centering ~ & Ours PPRB w/ fixed $\beta$ & 70.8 & 74.6 & 76.8 & 77.9 & 78.4 & 78.6 & 78.9 & 79.1 & 79.5 & 77.2 \\
  \centering ~ & Ours PPRB w/ RV & 71.0 & 74.4 & 76.3 & 77.1 & 78.2 & 79.0 & 79.4 & 79.6 & 80.1 & 77.2 \\
  \centering ~ & Ours w/ SA & 72.6 & 75.7 & 77.4 & 78.6 & 79.4 & 79.7 & 80.2 & 80.5 & 80.6 & 78.3 \\
  \centering ~ & Ours w/ RV-Mixup & 71.2 & 75.0 & 77.1 & 78.3 & 78.9 & 79.6 & 79.8 & 80.5 & 80.5 & 77.9 \\
  \centering ~ & Ours & 72.5 & 76.0 & 77.6 & 78.7 & 79.6 & 79.8 & 80.0 & 80.5 & 80.8 & 78.4 \\
  \hline
  \hline
  \centering \multirow{10}*{VG-200} & SSGRL & 34.6 & 37.3 & 39.2 & 40.1 & 40.4 & 41.0 & 41.3 & 41.6 & 42.1 & 39.7 \\
   \centering ~ & SSGRL w/ IP-Mixup & 33.6 & 36.9 & 38.9 & 40.1 & 40.6 & 41.4 & 41.8 & 41.4 & 41.9 & 39.6 \\
  \centering ~ & SSGRL w/ FM-Mixup & 34.3 & 36.7 & 39.4 & 40.3 & 40.7 & 41.8 & 41.4 & 42.2 & 42.6 & 39.9 \\
  \centering ~ & Ours IPRB & 39.0 & 42.0 & 44.0 & 44.6 & 44.9 & 45.6 & 46.2 & 46.6 & 47.2 & 44.5 \\
  \centering ~ & Ours IPRB w/ fixed $\alpha$ & 38.4 & 41.4 & 42.5 & 43.0 & 43.6 & 44.2 & 44.7 & 45.1 & 45.9 & 43.2 \\
  \centering ~ & Ours PPRB & 40.1 & 42.7 & 44.1 & 44.8 & 45.2 & 46.0 & 46.8 & 47.0 & 47.5 & 44.9 \\
  \centering ~ & Ours PPRB w/ fixed $\beta$ & 37.9 & 41.8 & 42.8 & 43.1 & 43.6 & 44.1 & 44.8 & 45.5 & 46.0 & 43.3 \\
  \centering ~ & Ours PPRB w/ RV & 40.6 & 43.3 & 44.2 & 44.6 & 45.2 & 46.0 & 46.4 & 46.8 & 47.4 & 44.9 \\
  \centering ~ & Ours w/ SA & 41.1 & 43.7 & 44.6 & 45.4 & 46.2 & 47.2 & 47.6 & 47.9 & 48.0 & 45.7 \\
  \centering ~ & Ours w/ RV-Mixup & 40.6 & 43.5 & 44.5 & 45.3 & 46.0 & 47.1 & 47.2 & 47.8 & 48.1 & 45.6 \\
  \centering ~ & Ours & 41.4 & 44.0 & 44.8 & 45.5 & 46.6 & 47.5 & 47.8 & 48.0 & 48.2 & 46.0 \\
  \hline
  \hline
  \centering \multirow{10}*{Pascal VOC 2007} & SSGRL & 77.7 & 87.6 & 89.9 & 90.7 & 91.4 & 91.8 & 91.9 & 92.2 & 92.2 & 89.5 \\
  \centering ~ & SSGRL w/ IP-Mixup & 75.0 & 86.8 & 86.3 & 90.4 & 91.1 & 91.3 & 91.3 & 91.7 & 92.0 & 88.4 \\
  \centering ~ & SSGRL w/ FM-Mixup & 78.8 & 88.3 & 88.9 & 91.0 & 91.5 & 91.6 & 91.6 & 91.8 & 92.4 & 89.5 \\
  \centering ~ & Ours IPRB & 84.1 & 87.7 & 90.6 & 91.6 & 92.2 & 92.6 & 92.7 & 92.9 & 93.0 & 90.8 \\
  \centering ~ & Ours IPRB w/ fixed $\alpha$ & 78.6 & 86.1 & 89.4 & 91.0 & 91.6 & 92.0 & 92.5 & 92.7 & 93.0 & 89.7 \\
  \centering ~ & Ours PPRB & 84.9 & 88.7 & 90.8 & 91.8 & 92.1 & 92.6 & 92.8 & 93.0 & 93.1 & 91.1 \\
  \centering ~ & Ours PPRB w/ fixed $\beta$ & 77.7 & 84.9 & 88.2 & 91.5 & 91.8 & 92.2 & 92.4 & 92.8 & 92.9 & 89.4 \\
  \centering ~ & Ours PPRB w/ RV & 81.7 & 88.1 & 90.1 & 91.4 & 91.8 & 92.1 & 92.5 & 92.6 & 92.8 & 90.3 \\
  \centering ~ & Ours w/ SA & 86.2 & 89.5 & 91.2 & 91.8 & 92.3 & 92.5 & 92.6 & 92.9 & 93.0 & 91.3 \\
  \centering ~ & Ours w/ RV-Mixup & 83.5 & 88.6 & 90.7 & 91.4 & 91.9 & 92.2 & 92.6 & 92.8 & 92.9 & 90.7 \\
  \centering ~ & Ours & 85.7 & 89.8 & 91.8 & 92.0 & 92.3 & 92.7 & 92.9 & 93.1 & 93.2 & 91.5 \\
  \hline      
  \end{tabular}
  \vspace{10pt}
  \caption{Comparison of mAP of the baseline SSGRL, SSGRL with mixup on image pixel level (IP-Mixup), SSGRL with mixup on feature map level (FM-Mixup), our DSRB merely using IPRB module (Our IPRB), our DSRB merely using IPRB module with fixed $\alpha$ (Ours IPRB w/ fixed $\alpha$), our DSRB merely using PPRB module (Ours PPRB), our DSRB merely using PPRB module with fixed $\beta$ (Ours PPRB w/ fixed $\beta$), our DSRB merely using PPRB module with category-specific representation vector level blending (Ours PPRB w/ RV), our DSRB with mixup on category-specific representation vector level (Ours w/ RV-Mixup), our DSRB using SA instead of SD (Ours w/ SA) and our DSRB (Ours) on the MS-COCO, VG-200 and Pascal VOC 2007 datasets. The best results are highlighted in bold.}
  \label{tab:ablation-results}
\end{table*}

\subsection{Comparison with the state-of-the-art algorithms}
To evaluate the effectiveness of the proposed DSRB, we compare it with the following algorithms that can be classified into three folds:  1) \textbf{SSGRL} (ICCV'19) \cite{Chen2019SSGRL}, \textbf{GCN-ML} (CVPR'19) \cite{Chen2019ML-GCN}, \textbf{P-GCN} (TPAMI'21) \cite{Chen2021P-GCN}, \textbf{KGGR} (TPAMI'22) \cite{Chen2022KGGR} achieve state-of-the-art performance on the traditional MLR task by exploring label co-occurrence or learning semantic-aware representation. For fair comparisons, we adapt these methods to address the MLR-PL task by replacing the BCE loss with partial BCE loss while keeping other components unchanged. However, since the code of P-GCN has not been released, we had to compare its provided results. 2) \textbf{CL} (CVPR'21) \cite{Durand2019CVPR} alternately labels the unknown labels with high evidence to update the training set and retrains the model with the updated training set. We also treat it as a strong baseline to address the MLR-PL task. 3) \textbf{Partial BCE} (CVPR'19) \cite{Durand2019CVPR}, \textbf{ASL} (ICCV'21) \cite{Ridnik2021ASL} are proposed to address MLR-PL and MLR task respectively by designing new loss function. The former introduces a normalized BCE loss to better exploit partial labels to train the multi-label models, and the latter introduces a novel asymmetric loss based on the focal loss \cite{lin2017focal-loss} to better balance negative and positive samples in the traditional multi-label image recognition task. For fair comparisons, we adopt the same ResNet-101 network as the backbone and follow exactly the same train/val split settings. 

\subsubsection{Performance on MS-COCO} 
We first present the performance comparisons on MS-COCO in Table \ref{tab:average-results} and \ref{tab:mAP-results}. As shown, our DSRB obtains the best performance over current state-of-the-art algorithms which achieves the average mAP, OF1, and CF1 of 78.4\%, 76.8\%, and 72.7\%, with an average improvement of 1.9\%, 2.8\%, 3.0\%. Besides these average results, we also present the mAPs on the all known label proportion settings for detailed analysis, as shown in Table \ref{tab:mAP-results}. Based on these results, we find CL algorithm suffers obvious performance degradation when the proportion of known labels is low (e.g., 10\%-30\%), which demonstrates current pseudo labels generating algorithms heavily rely on sufficient annotations for training. Orthogonal to these algorithms, our DSRB transfers information of known labels by blending category-specific semantic representations across different images. Thus, our DSRB performs consistently well on all known label proportion settings, even when the proportion of known labels is low. To be specific, it obtains the mAPs of 72.5\%, 76.0\%, 77.6\%, 78.7\%, 79.6\%, 79.8\%, 80.0\%, 80.5\%, 80.8\% on the settings of 10\%-90\% known labels, outperforming the second-best ASL algorithm by 2.8\%, 2.0\%, 2.5\%, 1.9\%, 2.1\%, 1.7\%, 1.3\%, 1.4\%, 1.1\%.

\subsubsection{Performance on VG-200}
Compared to MS-COCO, VG-200 is a more challenging benchmark dataset that covers much more categories. Thus, most algorithms achieve quite poor performance on this dataset. As presented in Table \ref{tab:average-results}, the previous best-performing KGGR algorithm obtains the average OF1, and CF1 of 41.2\%, and 33.6\% while our DSRB obtains the average OF1, and CF1 of 45.1\%, and 37.7\%. Similarly, our DSRB also obtains the best mAP performance which achieves the average mAP of 46.0\%, with an average improvement of 4.5\%. To attain an in-depth understanding, we present the mAP comparisons over different known proportion settings in Table \ref{tab:mAP-results}. As shown, the proposed DSRB outperforms current leading algorithms on all known label proportions. Concretely, it obtains the mAPs of 41.4\%, 44.0\%, 44.8\%, 45.5\%, 46.6\%, 47.5\%, 47.8\%, 48.0\%, 48.2\% on the settings of 10\%-90\% known labels, outperforming the second-best KGGR algorithm by 5.4\%, 4.0\%, 3.6\%, 4.0\%, 4.6\%, 5.0\%, 4.5\%, 4.4\%, 4.4\%.

\subsubsection{Performance on Pascal VOC 2007}
Pascal VOC 2007 is the most widely used dataset for evaluating multi-label image recognition, thus we present the performance comparisons on this dataset in Table \ref{tab:average-results} and \ref{tab:mAP-results}. As this dataset covers merely 20 categories, it is a much simpler dataset and current algorithms can also achieve quite well performance. However, our DSRB still achieves consistent performance improvement. As shown, it improves the average mAP, OF1, and CF1 by 1.0\%, 0.4\%, and 1.2\%. It is worth noting that our DSRB achieves more mAP improvement when using the fewer known labels. Specifically, it outperforms the second-best ASL algorithm by 2.8\%, 1.2\%, 1.8\%, when the known labels are merely 10\%, 20\% and 30\%.

\section{Ablative Studies} \label{sec:ablative}
In this section, we conduct comprehensive ablative studies to analyze the actual contributions of each crucial module in our DSRB to attain a better understanding. 

\subsection{Analysis of the CSRL module}
As discussed above, the CSRL module is introduced to extract category-specific semantic feature maps, and there are two current algorithms to implement this module, i.e., semantic decoupling (SD) \cite{Chen2019SSGRL} and semantic attention mechanism (SA) \cite{Ye2020ADD-GCN}. Thus, we conduct experiments to analyze the effect of using different CSRL implementation algorithms. As presented in Table \ref{tab:average-ablation-results} and \ref{tab:ablation-results}, we find that using these algorithms obtains comparable performance, which suggests the proposed DSRB can build on different CSRL implementation algorithms and does not depend on the semantic decoupling module. On the other hand, using the semantic decoupling module (denoted as ``Ours") obtains slightly better performance than using the semantic attention mechanism (denoted as ``Ours w/ SA"), with an average mAP improvement of 0.1\%, 0.3\%, and 0.2\% on the MS-COCO, VG-200, and Pascal VOC 2007 datasets. Thus, we chose the semantic decoupling algorithm to implement the CSRL module for all other experiments.

\subsection{Analysis of the DSRB}
As we utilize the semantic decoupling algorithm to implement the CSRL module and gated neural network for multi-label image recognition, SSGRL \cite{Chen2019SSGRL} can be regarded as the baseline. Here, we emphasize the comparisons with SSGRL to evaluate the effectiveness of DSRB as a whole module. As shown in Table \ref{tab:ablation-results}, the baseline SSGRL obtains the average mAP of 74.1\%, 39.7\%, and 89.5\% on the MS-COCO, VG-200, and Pascal VOC 2007 datasets. By introducing the DSRB to complement unknown labels, it improves the average mAP to 78.4\%, 46.0\% and 91.5\% on these three MLR-PL datasets, with the average mAP improvement of 4.3\%, 6.3\% and 2.0\%, respectively. Besides the average mAP, it also achieves similar improvement on the average OF1, CF1 metrics as presented in Table \ref{tab:average-ablation-results}.

Besides, the proposed DSRB, as a novel blending regularization, needs to compare with other blending regularization algorithms to demonstrate its superiority. Thus, we further conduct two baseline algorithms that perform position-wise blending in image pixel space and global feature maps space (namely ``SSGRL w/ IP-Mixup" and ``SSGRL w/ FM-Mixup"). As shown in Table \ref{tab:ablation-results}, both two baseline algorithms hard to improve the performance, i.e., ``SSGRL w/ IP-Mixup" obtains the average mAP of 74.2\%, 39.6\% and 88.4\%, while `SSGRL w/ FM-Mixup" obtains the average mAP of 74.4\%, 39.9\%, and 89.5\% on the three datasets, respectively. One likely reason for this phenomenon, as aforementioned in Section \ref{sec:rethinking}, is the MLR images generally interact with multiple objects, and simple blending regularization in these two spaces easily confuse semantic information among different categories and complex backgrounds which may degrade the performance. As presented in the preliminary version of this work \cite{Pu2022SARB}, performing blending regularization in category-specific representation vector space may alleviate the above problem, thus we conduct experiments that our DSRB perform blending in category-specific representation vector space (denoted as ``Ours w/ RV-Mixup"), and compare it with baseline SSGRL and our DSRB. As shown in Table \ref{tab:ablation-results}, it obviously outperforms the baseline SSGRL on all known label proportions but is worse than our DSRB. Concretely, it obtains an average mAP of 77.9\%, 45.6\% and 90.7\% on the MS-COCO, VG-200 and Pascal VOC 2007 dataset, with an degradation of 0.5\%, 0.4\% and 0.8\% compared with our DSRB (denoted as ``Ours"). One likely reason for this phenomenon is that blending samples in category-specific representation vector space ignores the localization information of objects, which easily generates confusing blended samples which may disturb the training process.

Considering the proposed DSRB consists of the instance-perspective representation blending module and the prototype-perspective representation blending module, we further conduct experiments to analyze these two modules to attain a better understanding of these modules.

\begin{figure*}[!t] 
  \centering    
  \subfloat{
  \includegraphics[width=0.3\linewidth]{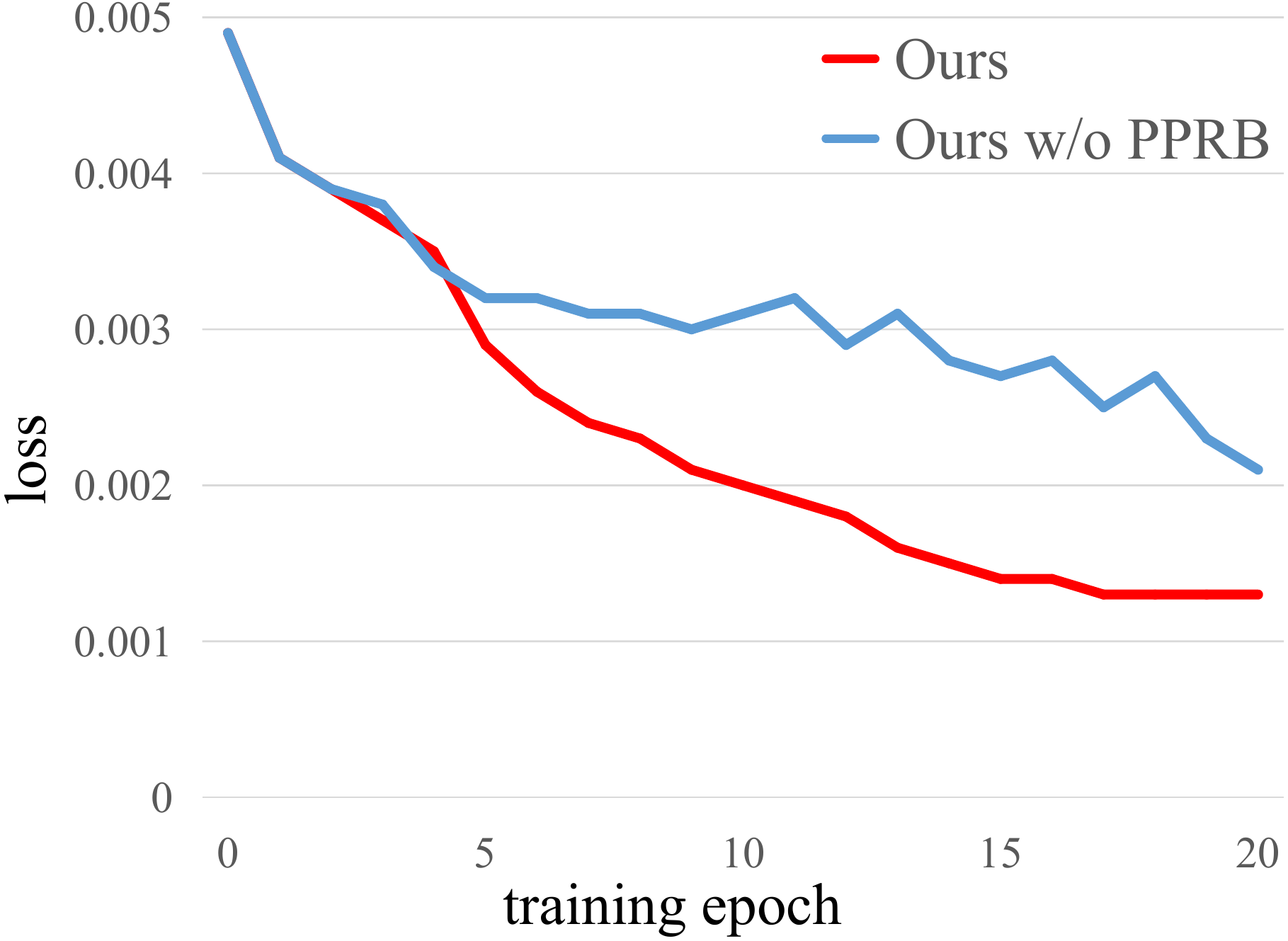}  
  }~     
  \subfloat{
  \includegraphics[width=0.3\linewidth]{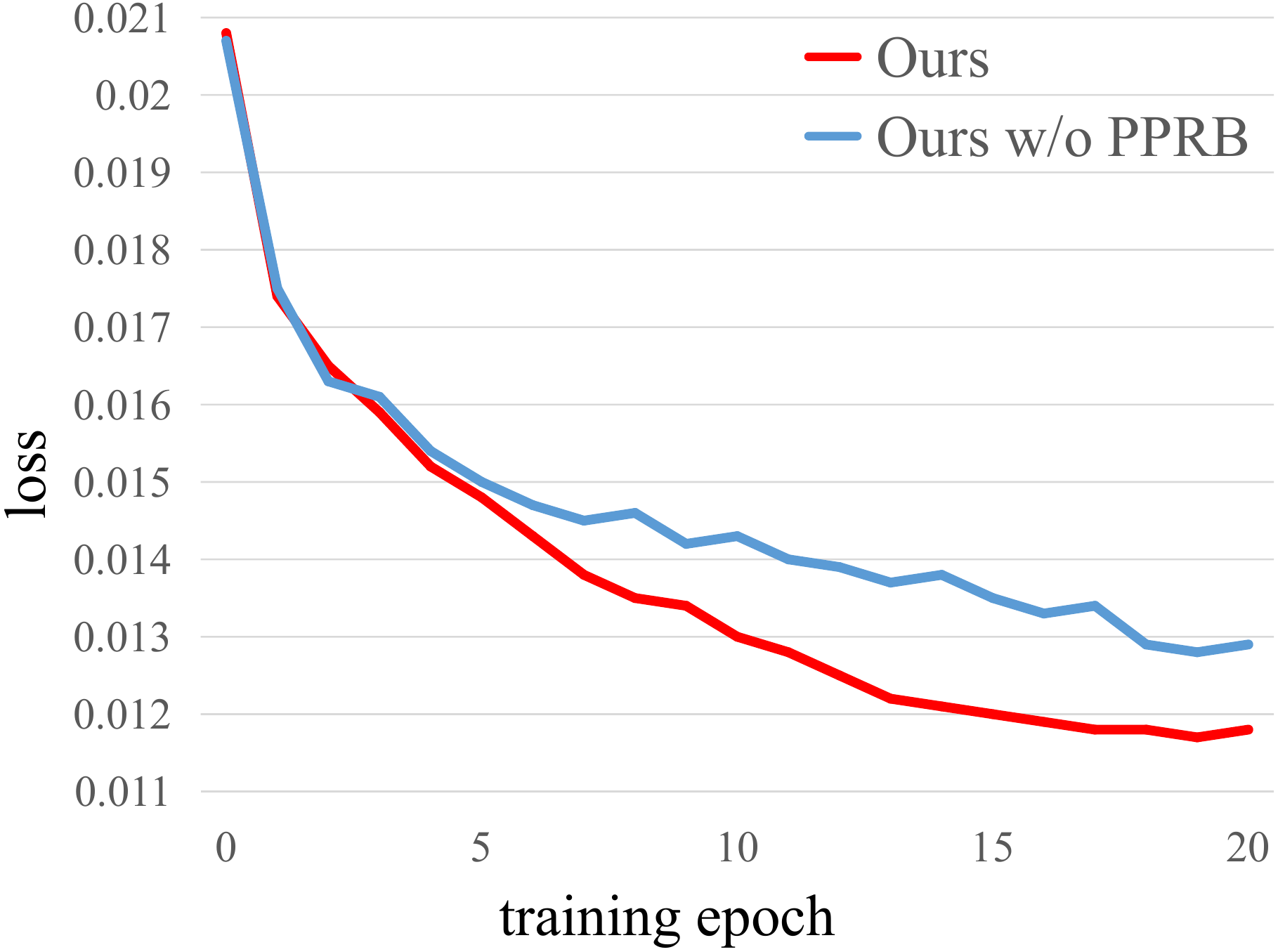}     
  }~
  \subfloat{
  \includegraphics[width=0.3\linewidth]{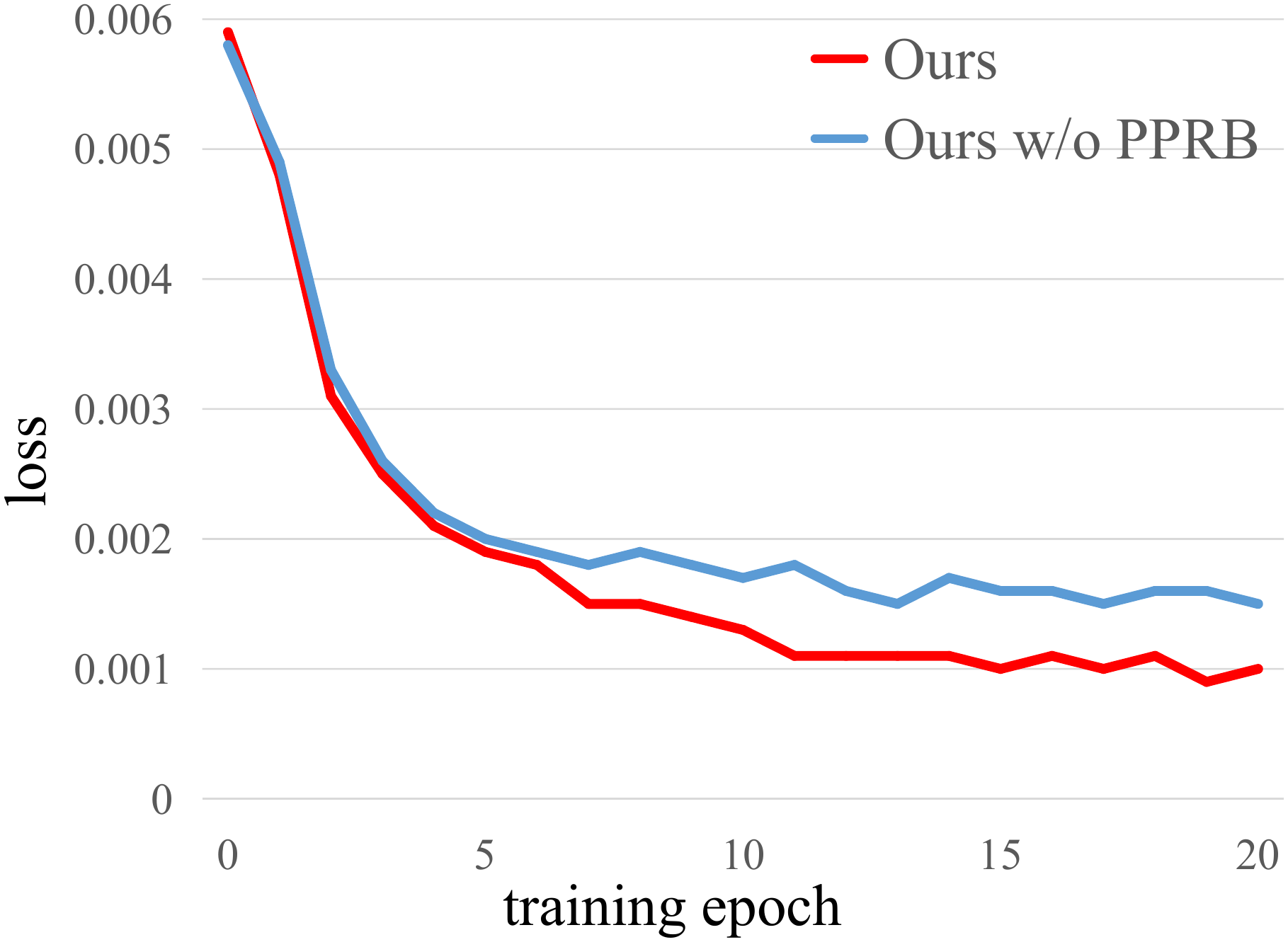}     
  }~
  \caption{Analysis of the effect of prototype-perspective representation blending module. These experiments are conducted on MS-COCO (left), VG-200 (middle) and Pascal VOC 2007 (right).}     
  \label{fig:loss-result}     
\end{figure*}

\subsection{Analysis of the IPRB module}
To analyze the actual contribution of the IPRB module, we conduct experiments that our DSRB merely uses this module (namely, ``Ours IPRB") and compare it with the SSGRL baseline on the MS-COCO, VG-200, Pascal VOC 2007 datasets. As shown in Table \ref{tab:ablation-results}, it obtains an average mAP of 77.8\%, 44.5\%, 90.8\% on MS-COCO, VG-200, Pascal VOC 2007, with the mAP improvement of 3.7\%, 4.8\%, and 1.3\% respectively. That results demonstrate the effectiveness of instance-perspective representation blending in the MLR-PL task.

The IPRB module contains a crucial parameter $\alpha$ that controls the ratio of instance-perspective semantic feature map blending. However, it is impractical and exhausting to find an optimal value for different datasets and known label proportion settings. In this work, we set $\alpha$ as a learnable parameter to adaptively learn the optimal value via standard back-propagation. To verify its contribution, we conduct experiments to compare with the IPRB module using a fixed $\alpha$ of 0.5 (namely ``Ours IPRB w/ fixed $\alpha$"). As shown in Table \ref{tab:ablation-results}, using a fixed value of 0.5 decreases the average mAPs from 77.8\%, 44.5\% and 90.8\% to 77.0\%, 43.2\% and 89.7\%, with the mAP degeneration of 0.8\%, 1.3\%, and 1.1\% respectively. 

\subsection{Analysis of the PPRB module}
Besides the IPRB module, the PPRB module is another important module that blends the representation of unknown labels with the prototypes of corresponding labels to complement these unknown labels. In this part, we analyze its effectiveness by comparing the performance with and without this module. 

As shown in Table \ref{tab:ablation-results}, adding the PPRB module to the baseline SSGRL (denoted as ``Ours PPRB") leads to 3.9\%, 5.2\%, and 1.6\% average mAP improvement on the MS-COCO, VG-200, and Pascal VOC 2007 datasets. As previously suggested, the PPRB module can help to avoid generating confusing training samples by introducing more robust representation prototypes, which make the training process be more stable. To validate this point, we further visualize the loss of the training process when the known label proportion is 50\%. As presented in Figure \ref{fig:loss-result}, it can be observed that the loss is choppy without the PPRB module, and adding this module can stabilize the training process. As aforementioned in Section \ref{sec:method}, the PPRB module can generate more stable blended samples by utilizing the coarse localization information in the category-specific semantic feature map space. To evaluate this point, we conduct experiments that our DSRB merely uses PPRB to perform blending in category-specific representation vector space (denoted as ``Ours PPRB w/ RV"). The experiment results show that removing the coarse localization information decreases the average mAPs from 78.0\%, 91.1\% to 77.2\%, 90.3\% on the MS-COCO and Pascal VOC 2007 datasets, with the mAP degeneration of 0.8\%, 0.8\% respectively. 

Similar to the IPRB module, the PPRB module also contains a crucial parameter $\beta$ that controls the ratio of prototype-perspective semantic feature map blending. To avoid searching an optimal value for different datasets and known label proportion settings, we also set $\beta$ as a learnable parameter to adaptively learn the optimal value via standard back-propagation. Here, we also conduct experiments to compare with the setting that fixed $\beta$ to 0.5 (namely ``Ours PPRB w/ fixed $\beta$") on the MS-COCO, VG-200, Pascal VOC 2007 datasets. As presented in Table \ref{tab:ablation-results}, it obtains an average mAP of 77.2\%, 43.3\%, 89.4\% on these three datasets, with the mAP degeneration of 0.8\%, 1.6\% and 1.7\% respectively. 

\section{Conclusion} \label{sec:conclusion}
In this work, we present a new perspective to complement the unknown labels by blending category-specific semantic representation to facilitate the MLR-PL task, which does not rely on sufficient annotations and thus can outperform current leading algorithms on all known label proportion settings. Specifically, the proposed DSRB consists of an IPRB module that blends representation of known labels to the representation of unknown labels that belong to the same category and from different images to generate diverse blended samples and a PPRB module that learns robust representation prototypes and generates more stable blended samples based on potential object presence areas. We conduct extensive experiments on multiple MLR-PL benchmark datasets (e.g., MS-COCO, VG-200, and Pascal VOC 2007) to demonstrate its superiority over current state-of-the-art algorithms and comprehensive ablative studies to analyze the actual contribution of each crucial module for in-depth understanding.

\bibliographystyle{IEEEtran}
\bibliography{reference}

\newpage

\begin{IEEEbiography}[{\includegraphics[width=1in,height=1.25in,clip,keepaspectratio]{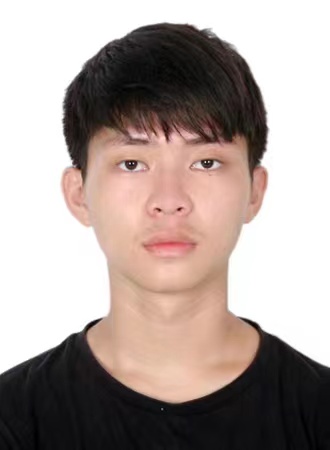}}]{Tao Pu} received a B.E. degree from the School of Computer Science and Engineering, Sun Yat-sen University, Guangzhou, China, in 2020, where he is currently pursuing a Ph.D. degree in computer science. He has authored and coauthored approximately 10 papers published in top-tier academic journals and conferences, including T-PAMI, AAAI, ACM MM, etc.\end{IEEEbiography}

\begin{IEEEbiography}[{\includegraphics[width=1in,height=1.25in,clip,keepaspectratio]{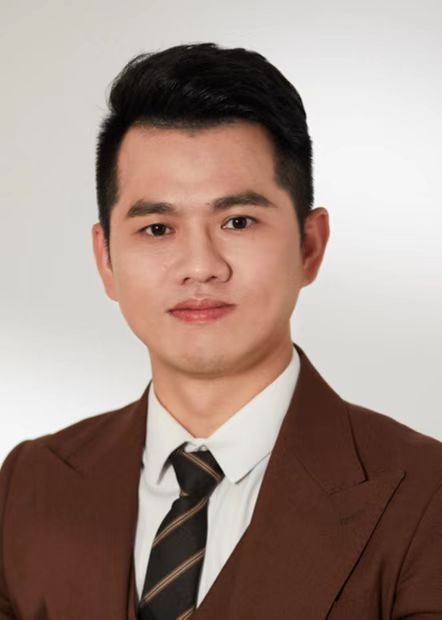}}]{Tianshui Chen} received a Ph.D. degree in computer science at the School of Data and Computer Science Sun Yat-sen University, Guangzhou, China, in 2018. Prior to earning his Ph.D, he received a B.E. degree from the School of Information and Science Technology in 2013. He is currently an associated professor in the Guangdong University of Technology. His current research interests include computer vision and machine learning. He has authored and coauthored approximately 40 papers published in top-tier academic journals and conferences, including T-PAMI, T-NNLS, T-IP, T-MM, CVPR, ICCV, AAAI, IJCAI, ACM MM, etc. He has served as a reviewer for numerous academic journals and conferences. He was the recipient of the Best Paper Diamond Award at IEEE ICME 2017. \end{IEEEbiography}

\begin{IEEEbiography}[{\includegraphics[width=1in,height=1.25in,clip,keepaspectratio]{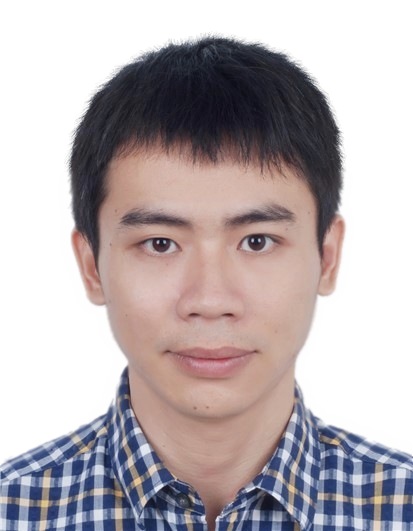}}]{Hefeng Wu} received a B.S. degree in computer science and technology and a Ph.D. degree in computer application technology from Sun Yat-sen University, China, in 2008 and 2013, respectively. He is currently a full research scientist at the School of Computer Science and Engineering, Sun Yat-sen University, China. His research interests include computer vision, multimedia, and machine learning. He has published works in and served as reviewers for many top-tier academic journals and conferences, including T-PAMI, T-IP, T-MM, CVPR, ICCV, AAAI, ACM MM, etc. \end{IEEEbiography}

\begin{IEEEbiography}[{\includegraphics[width=1in,height=1.25in,clip,keepaspectratio]{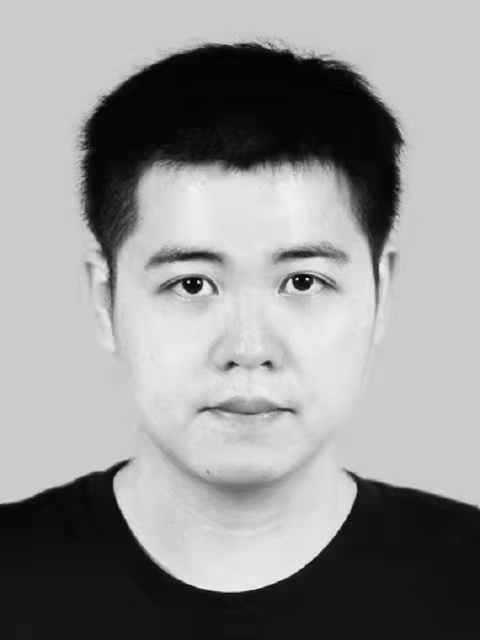}}]{Yukai Shi}
received a Ph.D. degrees from the School of Data and Computer Science, Sun Yat-sen University, Guangzhou China, in 2019. He is currently a lecturer at the School of Information Engineering, Guangdong University of Technology, China. His research interests include computer vision and machine learning.
\end{IEEEbiography}

\begin{IEEEbiography}[{\includegraphics[width=1in,height=1.25in,clip,keepaspectratio]{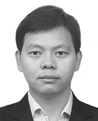}}]{Zhijing Yang} received B.S. and Ph.D. degrees from the School of Mathematics and Computing Science, Sun Yat-sen University, Guangzhou China, in 2003 and 2008, respectively. He was a Visiting Research Scholar at the School of Computing, Informatics and Media, University of Bradford, U.K., between July and December 2009, and a Research Fellow at the School of Engineering, University of Lincoln, U. K, between Jan. 2011 and Jan. 2013. He is currently a Professor and Vice Dean at the School of Information Engineering, Guangdong University of Technology, China. He has published over 80 peer-reviewed journal and conference papers, including papers for IEEE T-CSVT, T-MM, T-GRS, PR, etc. His research interests include machine learning and pattern recognition.
\end{IEEEbiography}

\begin{IEEEbiography}[{\includegraphics[width=1in,clip]{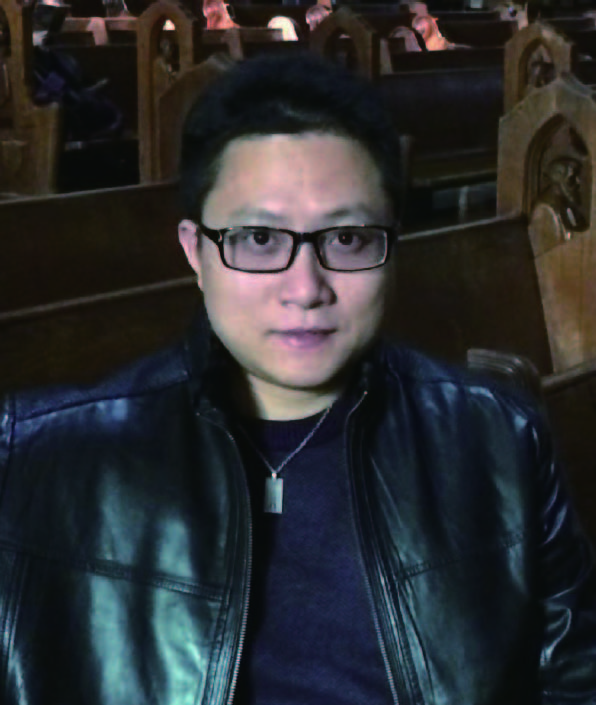}}]{Liang Lin} (M'09, SM'15) is a full professor at Sun Yat-sen University. From 2008 to 2010, he was a postdoctoral fellow at the University of California, Los Angeles. From 2016--2018, he led the SenseTime R\&D teams to develop cutting-edge and deliverable solutions for computer vision, data analysis and mining, and intelligent robotic systems. He has authored and coauthored more than 100 papers in top-tier academic journals and conferences (e.g., 15 papers in TPAMI and IJCV and 60+ papers in CVPR, ICCV, NIPS, and IJCAI). He has served as an associate editor of IEEE Trans. Human-Machine Systems, The Visual Computer, and Neurocomputing and as an area/session chair for numerous conferences, such as CVPR, ICME, ACCV, and ICMR. He was the recipient of the Annual Best Paper Award by Pattern Recognition (Elsevier) in 2018, the Best Paper Diamond Award at IEEE ICME 2017, the Best Paper Runner-Up Award at ACM NPAR 2010, Google Faculty Award in 2012, the Best Student Paper Award at IEEE ICME 2014, and the Hong Kong Scholars Award in 2014. He is an IET Fellow. \end{IEEEbiography}

\end{document}